\documentclass{article}
\pdfoutput=1 
\usepackage[utf8]{inputenc} %
\usepackage[T1]{fontenc}    %
\usepackage[dvipsnames]{xcolor}
\usepackage[pagebackref,colorlinks=true,linkcolor=BlueViolet,citecolor=BlueViolet,urlcolor=RedViolet]{hyperref}       %
\usepackage{url}            %

\usepackage{amsmath}
\usepackage{amssymb}
\usepackage{amsthm}
\usepackage{latexsym}
\usepackage{stackrel}

\usepackage{algorithm}
\usepackage{algpseudocode}

\usepackage{booktabs}       %
\usepackage{nicefrac}       %
\usepackage{multirow}
\usepackage{caption} 
\usepackage{xspace}
\usepackage{graphicx}
\usepackage{placeins}
\usepackage{cleveref}

\usepackage[page]{appendix} %
\PassOptionsToPackage{numbers, square ,compress}{natbib}

\usepackage{siunitx}

\usepackage[preprint]{neurips_2023}

\title{Adaptive Conformal Regression with\\Jackknife+ Rescaled Scores}
\author{%
Nicolas Deutschmann\\
IBM Research\\
\texttt{deu@zurich.ibm.com}\\
\And
Mattia Rigotti\\
IBM Research\\
\texttt{mrg@zurich.ibm.com}\\
\And
Mar\'ia Rodr\'iguez Mart\'inez\\
IBM Research\\
\texttt{mrm@zurich.ibm.com}\\
}

\begin{document}
\newtheoremstyle{thm}%
{1.em}%
{.5em}%
{\itshape}%
{}%
{\bfseries}%
{}%
{\newline}%
{}%

\newtheoremstyle{external}%
{1.em}%
{1.em}%
{\itshape}%
{}%
{\bfseries}%
{}%
{\newline}%
{\thmnote{#3}\thmname{ #1}}%

\theoremstyle{thm}
\newtheorem{definition}{Definition}

\newtheorem{theorem}{Theorem}
\newtheorem{proposition}[theorem]{Proposition}
\newtheorem{lemma}[theorem]{Lemma}

\theoremstyle{external}
\newtheorem{citethm}{Theorem}

\newcommand{\eqdef}{\ensuremath\stackrel{\text{def}}{:=}\;}
\newcommand{\mi}[1]{\ensuremath\textsc{MI}\left(#1\right)}
\newcommand{\omag}[1]{\ensuremath \!\cdot\!10^{#1}} %
\mathchardef\mhyphen="2D %

\newcommand{\jprs}{Ours\xspace}

\newcommand{\nth}{\textsuperscript{th}\xspace}

\captionsetup[table]{skip=1em}

\makeatletter
\renewcommand{\appendixtocname}{Supplementary Material}

\renewcommand{\appendixpagename}{\@title \\ --- \\ Supplementary Material}

\renewcommand{\appendixpage}{%
  \vbox{%
    \hsize\textwidth
    \linewidth\hsize
    \vskip 0.1in
    \@toptitlebar
    \centering
    {\LARGE\bf \appendixpagename \par}
    \@bottomtitlebar
    \vskip 0.1in \@minus 0.1in
  }
}
\makeatother

\bibliographystyle{unsrtnat}

\renewcommand\backrefxxx[3]{%
  \hyperlink{page.#1}{\textsuperscript{[$\uparrow$\,page\,#1]}}%
}

\renewcommand{\algorithmicrequire}{\textbf{Input:}}
\renewcommand{\algorithmicensure}{\textbf{Output:}}

\newcommand\blfootnote[1]{%
  \begingroup
  \renewcommand\thefootnote{}\footnote{#1}%
  \addtocounter{footnote}{-1}%
  \endgroup
}

\maketitle

\begin{abstract}
Conformal regression provides prediction intervals with global coverage guarantees, but often fails to capture local error distributions, leading to non-homogeneous coverage. We address this with a new adaptive method based on rescaling conformal scores with an estimate of local score distribution, inspired by the Jackknife+ method, which enables the use of calibration data in conformal scores without breaking calibration-test exchangeability. Our approach ensures formal global coverage guarantees and is supported by new theoretical results on local coverage, including an \textit{a posteriori} bound on any calibration score.
The strength of our approach lies in achieving local coverage without sacrificing calibration set size, improving the applicability of conformal prediction intervals in various settings.
As a result, our method provides prediction intervals that outperform previous methods, particularly in the low-data regime, making it especially relevant for real-world applications such as healthcare and biomedical domains where uncertainty needs to be quantified accurately despite low sample data.
\end{abstract}

\section{Introduction}
\label{sec:intro}
Conformal prediction (CP)~\citep{vovkMachineLearningApplicationsAlgorithmic1999,saundersTransductionConfidenceCredibility1999,vovkAlgorithmicLearningRandom2005a} provide a framework to perform rigorous post-hoc uncertainty quantification on machine learning predictions.
CP converts the predictions of a model into sets of predictions that can guarantee any desired expected coverage (the probability that the right answer is contained in the predicted set) with finite calibration data and no constraints on distributions.

There has been much recent work developing the original idea of CP to make it more computationally efficient~\citep{papadopoulosInductiveConfidenceMachines2002,leiConformalPredictionApproach2015}, generalize the hypotheses of the method~\citep{tibshiraniConformalPredictionCovariate2020,barberConformalPredictionExchangeability2022}, control Type I and Type II errors~\citep{fischConformalPredictionSets2022,angelopoulosConformalRiskControl2023}, and ensure different types of conditional validity~\citep{vovkConditionalValidityInductive2012,romanoClassificationValidAdaptive2020}. 
The original formulation of CP was best-suited for classification problems, and indeed the early work of \citet{leiDistributionFreePrediction2012} identified that a straightforward application of the formalism would predict constant-size intervals as prediction sets, highlighting the need for methods to make these prediction intervals (PI) dynamic.
The establishment of a theoretical basis for conformal regression (CR) by~\citet{leiDistributionFreePredictiveInference2017} was followed by multiple new approaches for CR that maintain the original guarantees of CP while also improving local behavior~\citep{romanoConformalizedQuantileRegression2019,guanConformalPredictionLocalization2020,guanLocalizedConformalPrediction2022,hanSplitLocalizedConformal2023,linLocallyValidDiscriminative2021}.

These approaches take three possible routes: early work focused on devising non-conformity scores that themselves capture some locally-dependent information~\citep{papadopoulosNormalizedNonconformityMeasures2008,leiDistributionFreePredictiveInference2017}, while a second wave of methods relied on non-dynamic CP applied on machine learning (ML) models that predict sets or intervals~\citep{romanoConformalizedQuantileRegression2019,feldmanImprovingConditionalCoverage2021}.
Finally, the seminal work of~\citet{guanConformalPredictionLocalization2020} on dynamically weighting the nonconformity score empirical distribution sparked a healthy influx of new methods~\citep{linLocallyValidDiscriminative2021,guanLocalizedConformalPrediction2022,hanSplitLocalizedConformal2023,amoukouAdaptiveConformalPrediction2023}.

Our work revisits the early approach of encoding local knowledge in the score function itself, in light of the general advances of the field of conformal predictions.
In particular, we consider the locally-rescaled absolute error score of~\citet{leiDistributionFreePredictiveInference2017} (\textsc{MADSplit}), which was formulated in the context of split-conformal predictions and therefore relies on the training error score distribution, often very different from test errors due to overfitting.
We propose a new approach to data splitting, combining both split-conformal and Jackknife+~\cite{barberPredictiveInferenceJackknife2020}, a recent method that modifies the jackknife to achieve prediction coverage guarantees.
This allows predictions with a single ML model, local error scale measures on the calibration data, while preserving coverage guarantees.
Our method does not have formal \textit{local} coverage guarantees, but we provide new bounds on local coverage based on the statistical dependence of the input data and the nonconformity scores which allow us to tune our score function to obtain close-to-optimal prediction intervals. We complement these theoretical results with detailed empirical comparison with existing methods on numerical, sequence, and image data, showing the benefits of our solution both in terms of data efficiency and tuning.
\subsection{Contributions}
\begin{itemize}
    \item We introduce a new method to rescale nonconformity scores by an \textit{in-distribution} local mean estimate, using a combination of the split-conformal and Jackknife+ schemes.
    \item We prove a global marginal coverage guarantee for our new approach.
    \item We prove a new bound on local coverage based on the mutual information between the non-conformity scores and the input data.
    \item We show how this bound combined to our scheme enables a new principled tuning of score localizers, leading to improved local coverage.
    \item We compare our method empirically with other approaches, and show that our method is more robust to low calibration statistics than approaches based on reweighted empirical score distributions, while avoiding the biases of the original \textsc{MADSplit} approach to rescaled scores.
\end{itemize}
\subsection{Related Work}
There are three main classes of approaches for providing adaptive prediction intervals (PI) for conformal regression.

\paragraph{Reweighted scores.}
The earliest efforts on adaptive CR proposes to rescale the usual nonconformity scores~\citep{papadopoulosNormalizedNonconformityMeasures2008,leiDistributionFreePredictiveInference2017} with local information. The most prominent approach is the \textsc{MADSplit} approach~\citep{leiDistributionFreePredictiveInference2017}, which uses $s(\hat{y}(X),y) = |\hat{y}(X)-y|/\hat\sigma(X)$, where $\sigma(X)$ is an estimate of the conditional score mean $\mathbb{E}[|\hat{y}(X)-y| | X]$.
Importantly in comparison with our proposed method, in \textsc{MADSplit} this estimate is performed on the training split, which preserves CP coverage guarantees, but yields poor performance when errors differ between training and test sets.

\paragraph{Model-based interval proposals.}
A second type of approach is based on choosing machine-learning models that themselves encode some notion of uncertainty. The canonical formulation is conformal quantile regression~\citep{romanoConformalizedQuantileRegression2019}, although other approaches based on predicting distribution parameters exist. These are applicable provided that the conditional label distribution $p(y|X)$ is slow-varying and wide enough that predicting conditional quantiles is a valid learning objective. Another limitation of these methods is that poor modelling performance immediately leads to poor PI.
This makes PI for difficult-to-predict examples the least trustworthy, despite often being the most important.

\paragraph{Reweighted score distributions.}
The most recent type of adaptive conformal regression methods is based on computing adaptive distortions of the non-conformity score empirical cumulative distribution function (ECDF), proposed originally by~\citet{guanConformalPredictionLocalization2020}. Given a test point, the ECDF is modified by giving a similarity-based weight to each calibration point, yielding an estimate of the test-point-conditional score ECDF. %
If the ECDF estimator is good, the result is perfect local coverage and adaptivity. Multiple variants have been proposed to optimize computational complexity or ECDF estimation, including \textsc{LCR}~\citep{guanConformalPredictionLocalization2020}, \textsc{LVD}~\citep{linLocallyValidDiscriminative2021}, \textsc{SLCP}~\citep{hanSplitLocalizedConformal2023}. In any case, estimating the high quantiles of the conditional CDF can require significant statistics, as many calibration points are needed \textit{near every test point}. Furthermore, as we discuss in~\cref{sec:experiments}, the proposed localization-weight tuning methods often yield poor results on complex, high-dimensional data.

\subsection{Definitions}
\label{sec:defs}
Let us establish definitions that will be reused throughout this manuscript. We consider a probability space $M_{X\times y}=\left(\Omega_{X}\times\Omega_{y}, \mathcal{F}_{X}\otimes\mathcal{F}_{y}, \pi\right)$ defined over a product of measurable spaces\footnote{It goes without saying that the probability distribution $\pi$ is not a product measure, \textit{i.e.} $y\not\perp X$.}, $\left(\Omega_{X}, \mathcal{F}_{X}\right)$, and $\left(\Omega_{y}, \mathcal{F}_{y}\right)$, to which we refer respectively as \textit{input-} and \textit{label-space}.

We work in the context of split-conformal predictions, \textit{i.e.,} we sample a training dataset which we use to produce a predictive model $m(X): \Omega_X\to\Omega_y$ through a training algorithm. We then further sample \textit{i.i.d} $(X_i,y_i)_{i=1\ldots N+1}$, also independent from the training data and $m$. We refer to the first $N$ sampled points as the calibration dataset $\mathcal{C}_N = (X_i,y_i)_{i=1\ldots N}$ and to $(X_{N+1}, y_{N+1})$ as the test point. For conformal predictions, we consider a score function $s(X,y):\Omega_X\times\Omega_y\to\mathbb{R}$ and define the usual calibration intervals with marginal risk $\alpha\in [0,1]$ as $S^\alpha(X) = \left\{y\in\Omega_y| s(X,y) \leq q^{1-\alpha}_{\mathcal{C}_N}(s)  \right\} $, where $q^{1-\alpha}_{\mathcal{C}_N}(s)$ is the $\lceil (1-\alpha)(n+1) \rceil /n$-quantile of the empirical score distribution on $\mathcal{C}_N$.

\section{Local Coverage and Score-Input Independence}
\label{sec:local_coverage_independence}
We begin by defining and characterizing the goals of adaptive conformal regression in terms of guarantees and tightness. This analysis is general and extends beyond the method we propose, but will also provide motivation for our approach and a framework to understand its applicability.

Informally, the objective of PI prediction to achieve two desirable properties: local coverage guarantees for any input point $X$ and interval bounds that perfectly adapt to capture the label distribution.

\subsection{Local Coverage Guarantees}
We propose a slightly extended notion of conditional coverage compared to that of~\citet{hanSplitLocalizedConformal2023}, which is more suitable to proving the bound of~\cref{thm:weak_conditional_cov}.
\begin{definition}[Input-space strong conditional coverage (\textsc{iscc})]
\label{def:islcc}
We define $\alpha$-\textbf{input-space strong conditional coverage} as the following property:
\begin{equation*}
\forall \omega_{X} \in \mathcal{F}_{X},\, \mathbb{P}_{X,y\sim \pi}\left(y\in S^{(\alpha)}\left(X\right) \Big| X\in \omega_{X}\right) \geq 1-\alpha.
\end{equation*}
\end{definition}

The formulation of~\cref{def:islcc} strongly hints at as sufficient condition to ensure strong conditional coverage for conformal intervals defined as in \cref{sec:defs}. Weaker conditional coverage properties can be defined as the same condition applying to specific subsets of $\mathcal{F}_X$.

\begin{proposition}[Sufficient condition for \textsc{iscc}]
\label{thm:islclemma_main}
If $S^{(\alpha)}(X)$ is defined from a score $s(X,y)$,
\begin{equation*}
X \perp s\left(X,y\right) \Rightarrow \alpha\mhyphen\textsc{iscc}
\end{equation*}
\end{proposition}

Indeed, if the score distribution is input-independent, then its conditional quantiles are as well, so that estimating quantiles globally is the same as estimating them locally. This has been observed in previous work and is tightly related to the definition of the orthogonal loss in~\citet{feldmanImprovingConditionalCoverage2021}, which measures the correlation between the local coverage and the local interval size.

Score-input independence is never realized in practice, and we are not aware of any coverage bound based on the orthogonal loss when optimality is not achieved. As we show below, however, mutual information can be used instead to place a bound on local coverage.

\begin{proposition}[Bound on conditional coverage]
If the mutual information between $X$ and $s(X,y)$, $\textsc{MI}(X,s)$ is finite, for any $\omega_X\in\mathcal{F}_X$ such that \(\mathbb{P}\left(X\in\omega_{X}\right)>\rho\)
\begin{equation*}
\mathbb{P}\left( y \in S^{\alpha}(X) | X\in \omega_{X}  \right) \geq \left(1-\tilde{\alpha}\right),\text{ where } \tilde\alpha = \alpha + \frac{\displaystyle\sqrt{1-e^{-\textsc{MI}(X,s)}}}{\rho}.
\end{equation*}
\label{thm:weak_conditional_cov}
\end{proposition}

Note that this bound applies the coverage probability marginalized over the calibration data. The actual coverage given a calibration dataset is a random variable which can deviate below the bound significantly if the calibration sample size is limited~\citep{vovkConditionalValidityInductive2012}.

This results is a direct consequence of \cref{thm:weak_conditional_cov}, and is proven in the supplementary material. The bound is vacuous for low-probability sets, which, as we argue in~\cref{app:scoreinputindep}, is likely inherent to this type of result. On the other hand, given a score function with free parameters, one can extend its usefulness by optimizing for smaller score-input mutual information, as we illustrate in~\cref{app:kernel_tuning}.

\subsection{Measuring the Adaptivity of PI}
\label{sec:eval_metrics}
It is common to evaluate adaptive CR methods based on two metrics computed on held-out test data: the coverage rate (Cov.) and the mean PI size (IS), under the understanding that, at fixed coverage, smaller intervals means tighter correlation between PI size and error rate. However, as we show in \cref{app:mis_limits}, an oracle predicting ideal intervals might actually mean \textit{increasing} the mean interval size compared to a non-adaptive approach.
We therefore propose several new metrics to evaluate the adaptivity of a given PI prediction method, using absolute errors as our conformal score (CS):
\begin{itemize}
    \item $\tau_\text{SI}$, the point-wise Kendall rank correlation coefficient between the CS and the PI size (IS)%
    \item $\tau_\text{SQI}$, the Kendall correlation between IS quantiles (ISQ, we usually use deciles) and the ISQ-conditional score $(1-\alpha)$\textsuperscript{th} quantile (CSQ)
    \item $R^2_\text{SQI}$, the $R^2$ regression coefficient for the linear model $\text{ISQ}=2\times \text{CSQ}$, where we use the middle point of each IS quantile bin as value for $\text{ISQ}$\footnote{Note that Kendall's $\tau$ is rank-based and therefore the rank of each quantile can be used instead of a numerical value for $\tau_\text{SQI}$ and $\tau_\text{SAI}$.}
\end{itemize}
When the score is the absolute error, $\text{IS} = 2 \times q^{(1-\alpha)}\left(\text{CS} | \text{IS}\right)$ is a necessary condition for reaching the best possible interval-local error coupling. We propose $R^2_\text{SQI}$ to measure the validity of this relation, which relies on discretizing the interval size distribution to measure conditional error quantiles.

While nearly-perfect solutions can be compared with $R^2_\text{SQI}$, it treats adaptive but over-conservative PI on the same footing as PI with no adaptivity. We therefore also use the rank correlation coefficient $\tau$ as a measure of monotonicity, using point-level metrics in $\tau_\text{SI}$, and quantile-aggregated metrics in $\tau_\text{SQI}$, which is less sensitive to distribution shape details. Informally, these two metrics can be thought of as framing the orthogonal loss of~\citet{feldmanImprovingConditionalCoverage2021} in terms of how stringent the score-interval size relationship is constrained. Our $\tau_{SQI}$ can also be seen as a different compromise in granularity: we recover continuous information about coverage by using the conditional score quantiles but discretize the interval sizes.

\section{Theoretical Results: Jackknife+ Rescaled Conformal Scores}
\subsection{Locally Rescaled Conformity Metric}
\label{sec:conformity_metric}
Considering the observations made in \cref{sec:local_coverage_independence}, one can note that, for a given $\alpha$, we can achieve \textsc{iscc} by rescaling score function $s$:
\begin{equation}
    \sigma_{q_\alpha}(X,y) = \frac{s(X,y)}{q^{(1-\alpha)}(s|X)},
\label{eq:rescaled_quantile}
\end{equation}
where $q^{(1-\alpha)}(s|X)$ is the $(1-\alpha)$-th quantile of the conditional score distribution $p(s(X,y)|X)$, which is essentially what ECDF-based methods aim to do. It is however costly to estimate these conditional quantiles and we propose to revisit the generalisation of the rescale absolute error used in \textsc{MADSplit}~\citep{leiDistributionFreePredictiveInference2017}

\begin{equation}
    \sigma(X,y) = \frac{s(X,y)}{\hat{\bar{s}}(X)},
    \label{eq:ideal_jplus_score}
\end{equation}

where $\hat{\bar{s}}(X)$ is an estimator of the mean of $s$ conditioned on $X$, $\bar{s}(X)$. It is easy to show that this score yields again the perfect results of~\cref{eq:rescaled_quantile} when the random variable $s(X,y)|X$ is a scale family indexed by $X$. Empirically, it is also often the case that this ratio has reduced $X$-dependence compared to the raw score $s$.

The conditional mean can be estimated with a Nadaraya–Watson estimator based on a kernel $K$:

\begin{equation}
\hat{\bar{s}}_i = \sum_{j=1;j\neq i}^{N} p^{(K)}_{ij} s_j, \quad\text{ where } p_{ij}^{(K)} = \frac{K(X_i,X_j)}{\sum\limits_{k\leq N;k\neq i} K(X_i,X_k)}.
\label{eq:kernel_mean}
\end{equation}

Formal guarantees on the convergence of this estimator typically require fine-grained knowledge of the data distribution to be made quantitative, which is rarely the case in practice. Nevertheless, \cref{thm:weak_conditional_cov} provides a tool to guide kernel choice and tuning by minimizing $\textsc{MI}(X,s)$.

In \textsc{MADSplit}, $\hat{\bar{s}}_i$ is estimated on the training data, which is risky due to the training-test error distribution shift, especially for modern deep learning~\citep{powerGrokkingGeneralizationOverfitting2022}. If we were to instead use the calibration data as in~\cref{eq:kernel_mean}, we would predict the following test interval

\begin{equation}
C_{N+1} = \mu(X_{N+1}) \pm \hat{\bar{s}}_{N+1} q^{\alpha}\left( \left\{ \sigma_{i} \right\}_{i\in\text{cal}} \right).
\end{equation}

This, however, breaks the exchangeability of the test point with the calibration set, which breaks the hypotheses of split-CP.
Further splitting the calibration dataset is another possible option, but this requires sacrificing the variance of the calibration-conditional coverage~\citep{vovkConditionalValidityInductive2012}. Our new method described below uses a modification of~\cref{eq:kernel_mean} to restore formal guarantees.

\subsection{Exchangeable Rescaled Scores with Jackknife+}

We propose an approach that combines elements of the split conformal and Jackknife+ approaches~\citep{barberPredictiveInferenceJackknife2020} by splitting the data into a training and calibration set, and applying the Jackknife+ on the calibration set to train an estimator of the conditional conformal score mean.

After training a model, we consider a matrix of scores, indexed by the union of the calibration set and the test point, $J^+=[1\ldots N+1]$. Each score itself is evaluated by computing every estimator over a restricted set $J^+_{ij} = J^+/\left\{i,j\right\}$:
\begin{equation}
\hat{\bar{s}}_{ij} = \sum_{k\in J^+_{ij}} p^{(K)}_{ik;j} s_k, \quad\text{ where } p_{ik;j}^{(K)} = \frac{K_{ij}(X_i,X_j)}{\sum\limits_{l\in J^+_{ij}} K_{ij}(X_i,X_l)}.
\label{eq:kernel_mean_jplus}
\end{equation}
We introduced the kernels $K_{ij}$ to allow kernel tuning: all $K_{ij}$ can be taken from the same family of kernels and have their parameters optimized to minimize the test-score mutual information $\mi{s, X}$ estimated on $J^+_{ij}$. Another option is to fix a kernel \textit{a-priori}, which the robustness of the mean-rescaled score easily allows.

We can express the scores as $s_{ij}^{+} = \frac{\left|\mu(X_{i}) - Y_{i} \right|}{\hat{\bar{s}}_{ij}}$, and in particular the actual calibration scores are $s^+_i = s^+_{i(N+1)}$. These $(N+1)\times (N+1)$ scores are truly exchangeable over the whole calibration set extended with the test point. There are now $N$ scores for the test point, which get folded into the interval definition as follows:

\begin{equation}
C^{+\alpha}_{N+1} = \left[ \mu_{X_{N+1}} - q^{\alpha}\left( \left\{\hat{\bar{s}}_{N+1,i}s_{i}^{+} \right\} \right),~ \mu_{X_{N+1}} +q^{\alpha}\left( \left\{ \hat{\bar{s}}_{N+1,i}s_{i}^{+} \right\} \right)  \right].
\label{eq:jplus_interval}
\end{equation}

We provide a summary of our procedure in~\cref{alg:jplrs}

\subsection{Global Coverage Guarantee}
\label{sec:global_guarantee}
When using the conformal prediction intervals defined in~\cref{eq:jplus_interval}, we obtain exactly the same type of coverage guarantee as the original Jackknife+ approach provides:
\begin{proposition}[Global Coverage of $C^{+\alpha}_{N+1}$]
Marginalizing over the calibration set:
\begin{equation*}
\mathbb{P}\left(Y_{N+1} \in C^{+\alpha}_{N+1}(X_{N+1})\right) \geq 1-2\alpha.
\end{equation*}
\end{proposition}

This coverage is degraded compared to the coverage probability of standard conformal predictions. Nevertheless, as for the original Jackknife+ approach, we observe empirically that the effective coverage is most often close to $(1-\alpha)$. As for~\cref{thm:micov}, one should keep in mind that this bound is marginalized over calibration data.

\section{Experimental Evaluation}
\label{sec:experiments}
Throughout this section, unless mentioned otherwise, we do not use the kernel tuning capabilities of our approach, and fix the kernel to a simple (approximate~\cite{PyNNDescentFastApproximate,dongEfficientKnearestNeighbor2011}) K-nearest-neighbor (KNN) kernel ($K_{ij}=1$ if $j$ is among the KNN of $i$, otherwise $K_{ij}=0$), setting $K=10$. We've found that this approach yields competitive results in many cases without any tuning for our approach. We discuss the potential gains of kernel tuning in~\cref{app:kernel_tuning}.

\subsection{Comparing Fairly to Previous Work}
\label{sec:kernel_comparison}
Tuning a localizer is a key element of all existing post-hoc adaptive PI and should, to some extent, be considered part of the method. In particular, our proposal is, to our knowledge, the only one using training-independent score information without sacrificing $\mathcal{O}(N_\text{cal})$ data, and cannot be applied to the baselines to which we compare. To perform fair comparisons, we have tried to strike a balance between considering PI proposal methods as end-to-end and evaluating the CP method in isolation by making best effort extensions.

We evaluate our method against \textsc{MADSplit}~\citep{leiDistributionFreePredictiveInference2017} and \textsc{LVD}~\citep{linLocallyValidDiscriminative2021} as representatives of the two classes of post-hoc CR. We have indeed found that their proposed methods for kernel tuning lead to poor performance in our experiments, which we report in~\cref{app:more_results}. We use modified kernels that yield better performance and therefore help assess the CR methods themselves. For \textsc{MADSplit}, we use the same KNN kernel as for our method, but these are unsuited for \textsc{LVD}. The original kernel for \textsc{LVD} is an anisotropic RBF  ($K(x,y)=e^{-|A(x-y)|^2}$) optimized on training data. We found that the matrix $A$ is usually too large, yielding mostly infinite intervals. We therefore replace it by $\lambda A$, tuned on subsampled training data to have at least $(1-\alpha)$ samples with effective sample size\footnote{Measured as $\exp H(w)$ where $w$ is the kernel-reweighted MDF on the calibration set.} at least $2/\alpha$, allowing mostly finite intervals. For metrics, we replace infinite interval sizes by the largest observed calibration error. Note that we also attempted to evaluate~\textsc{SLCP}~\citep{hanSplitLocalizedConformal2023}, but were unable to tune kernels to reach comparable metrics to the other methods, and we therefore left it out of our results.

\subsection{One-Dimensional Regression}
\label{sec:1d_regression}
\begin{figure}[ht]
    \centering
    \includegraphics[width=\textwidth, trim={1cm 0 1cm 0},clip]{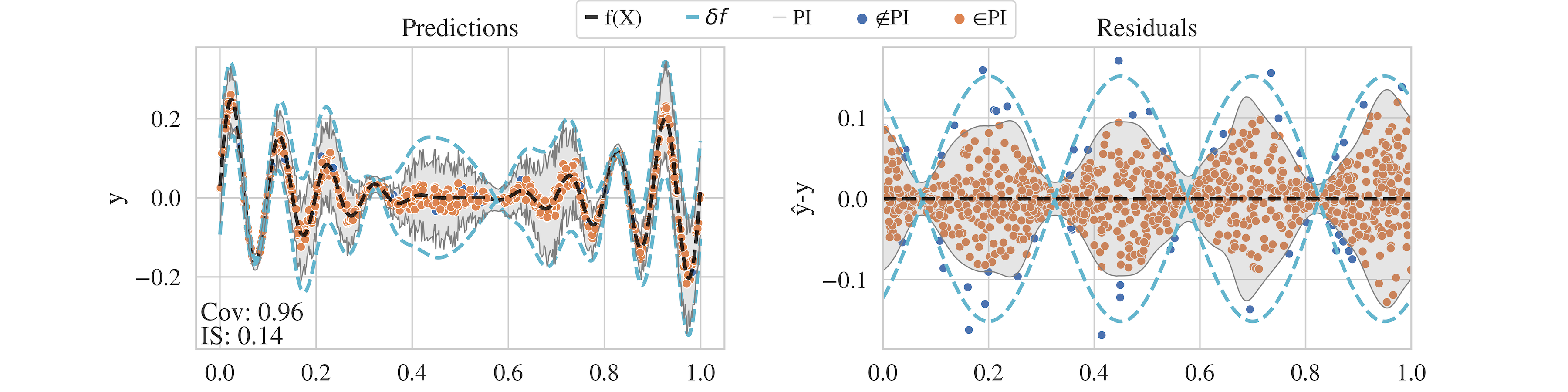}
    \caption{Evaluation of our approach on a 1D regression problem.}
    \label{fig:1d_fit}
\end{figure}

Let us start by applying our methods to simple regression problems that will help illustrate how our approach operates. We define the following random variables
\begin{equation}
\label{eq:1d_setup}
\def\arraystretch{2}
\begin{array}{c}
    X\sim \text{U}(0,1), \qquad y = f(X) + \epsilon, \qquad \epsilon  \sim \mathcal{N}\left(0, \nu(X) \right),\\
    f(X) = f_0 + X^2 \sin\left( k_f X + \phi_f \right), \qquad \nu(X) = \epsilon_0 + \left|\sin\left(k_\epsilon X+ \phi_\epsilon\right)\right|
\end{array}
\end{equation}
We use 1000 calibration and 10000 test points to evaluate our method on a random-forest regressor trained on independent data and indeed find intervals that achieve the target global coverage and dynamic interval sizes, as show in \cref{fig:1d_fit}.

\begin{figure}[ht]
    \centering
    \includegraphics[width=0.49\textwidth]{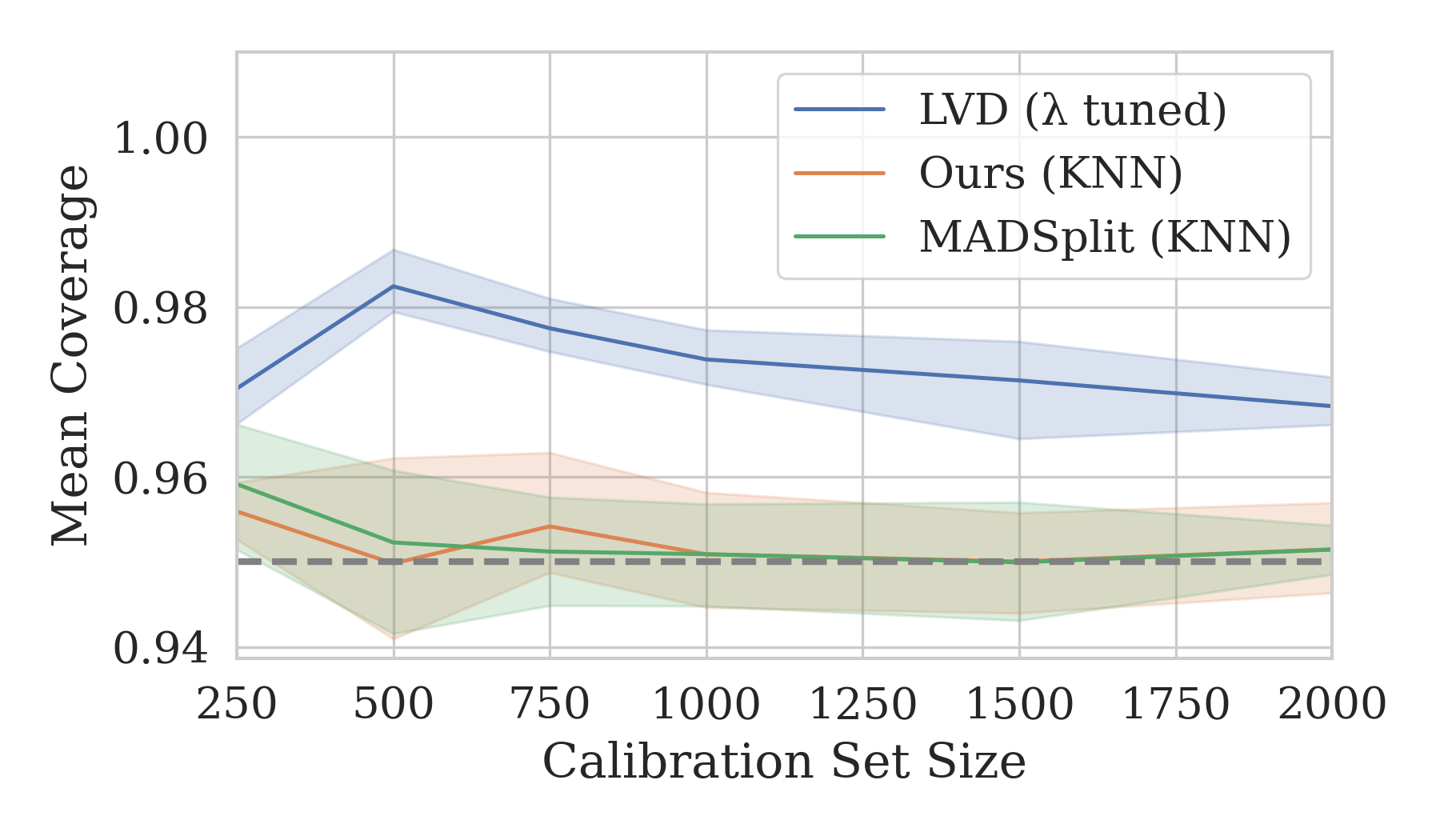}
    \includegraphics[width=0.49\textwidth]{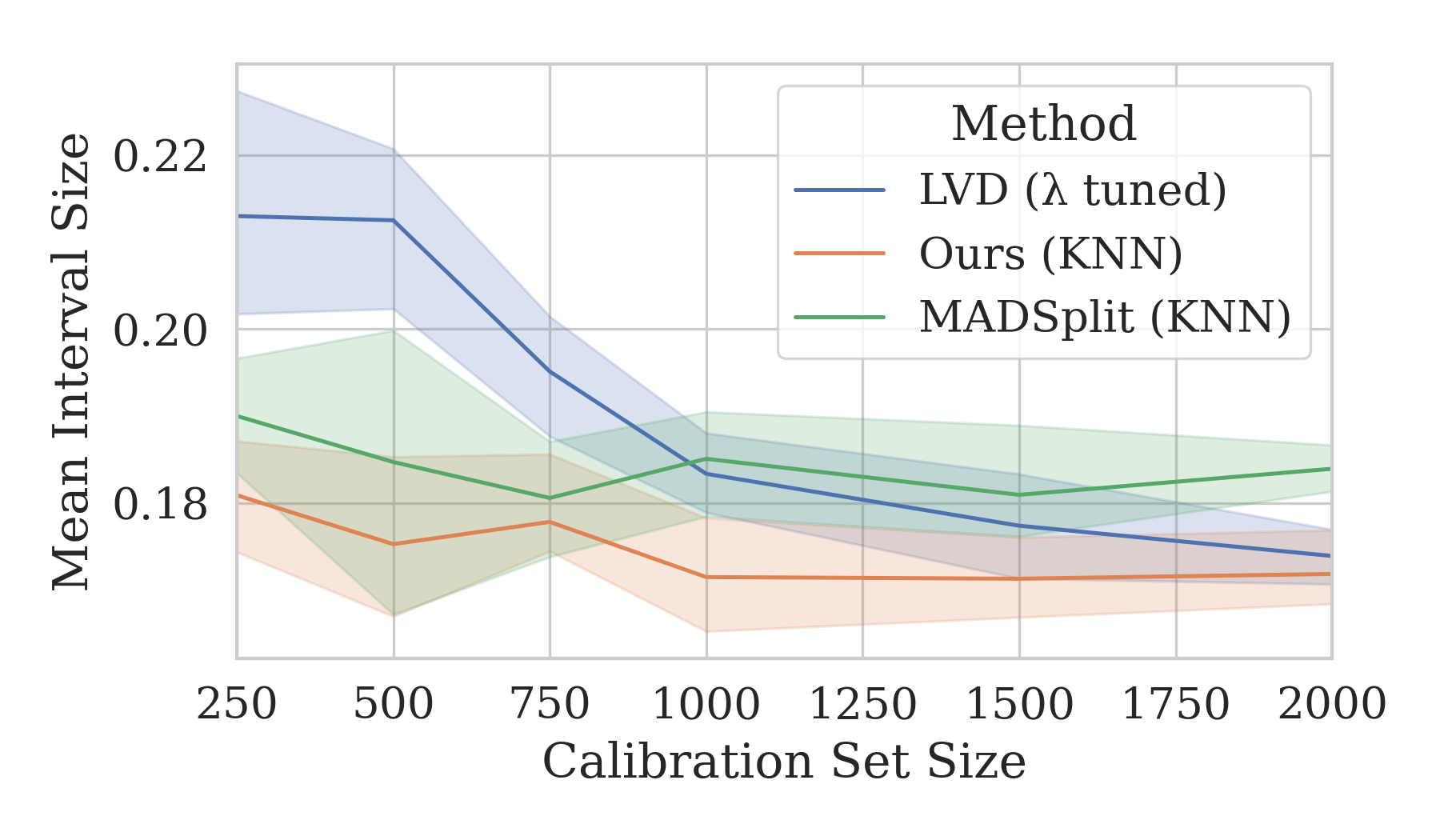}\\
        
    \includegraphics[width=0.49\textwidth]{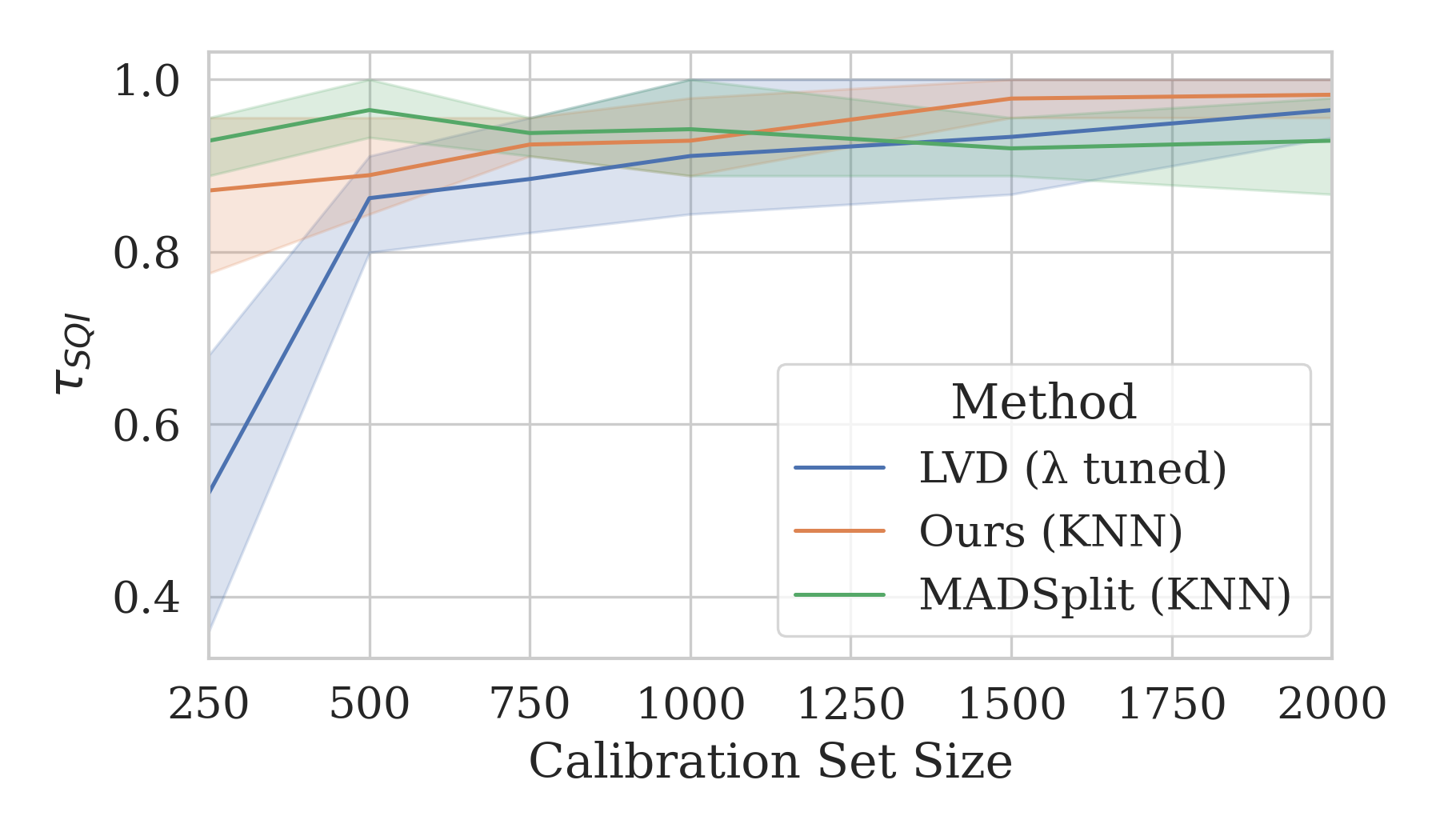}
    \includegraphics[width=0.49\textwidth]{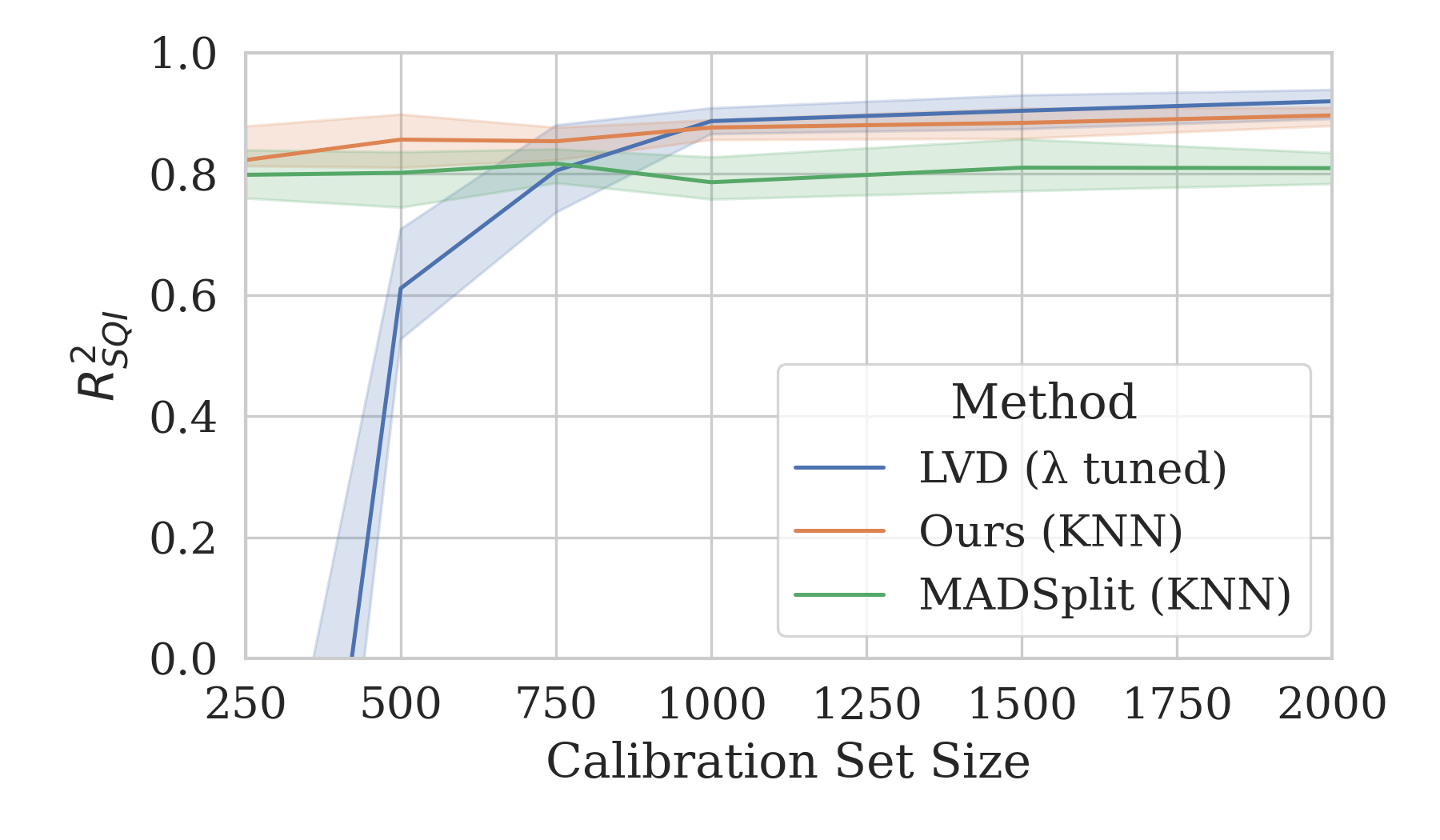}    
    \caption{Calibration-set-size dependence of PI metrics evaluated on an independent test set.}
    \label{fig:metrics_1d}
\end{figure}

This simple setting already allows us to compare performance with other adaptive PI methods. We display the calibration-set-size dependence of PI metrics in~\cref{fig:metrics_1d}.

These results highlight the case we make more systematically below: in the low data regime, our method is more robust than ECDF-based methods such as \textsc{LVD} due to its better data efficiency, and has comparable performance with them when more data is available. In this example, there is no significant difference with \textsc{MADSplit} in the low-data regime, while we outperform it in the high data regime, which we attribute to the training-test error distribution shift.

\subsubsection{Results on complex data}
\label{sec:exp_many_datasets}
In order to assess and compare the performance of our approach and previous work, we selected a number of regression task spanning data modalities and model performances. We provide results in the high- and low-calibration data regimes and compute the metrics defined in~\cref{sec:eval_metrics}. All experiments are performed with a target coverage of $1-\alpha=95\%$, repeating over 10 splits of the held-out data between calibration and test sets.

We consider four datasets in addition to our 1D regression problem. In all cases, we fix the training data and train an adapted regression model by optimizing the mean squared error. Details about data processing, embeddings used for localisation and predictive models can be found in~\cref{sec:reproducibility}. The tasks on which we evaluate are:

\textbf{Two \textsc{TDC} molecule property datasets~\citep{huangTherapeuticsDataCommons2021a}:} drug solubility and clearance prediction from respectively 4.2k and 1.1k \textsc{SMILES} sequences\footnote{string-based description of molecular structures~\citep{weiningerSMILESChemicalLanguage1988}.}.\\[.5em]
\textbf{The \textsc{AlphaSeq} Antibody binding dataset~\citep{engelhartDatasetComprisedBinding2022}:} binding score prediction for a SARS-CoV-2 target peptide from 71k amino-acid sequences.\\[.5em]
\textbf{Regression-MNIST~\citep{lecunMnistDatabaseHandwritten2005}:} Floating-point prediction of the label of MNIST images with test-time augmentations. This task was selected due to its irregular error distribution, which is expected to challenge our rescaling approach, see~\cref{app:mnist_details}.\\

We detail our results in~\cref{tab:big_results}. In summary, we confirm our results that our method outperforms \textsc{LVD} when data is scarce, while we are usually comparable with \textsc{MADSplit}, and that our performance improves faster than \textsc{MADSplits} with statistics. The results on sequence data with higher statistics highlight confirm that our method is competitive with both method. Interestingly, while AlphaSeq has the most data, our method dominates others. This is due to a combination of increased overfitting and scattered data distribution as discussed in~\cref{app:alphaseq_details}. Finally, the MNIST task is peculiar: all methods have comparable performance as measured by correlation measures but only \textsc{LVD} captures anywhere close to the ideal absolute interval size, as measured by $R^2_{SQI}$. We attribute this to the irregular error distributions due to possible number confusions, which leads to highly variable conditional mean/quantile ratios, as shown in~\cref{app:mnist_details}.

\begin{table}
    \centering
    \makebox[\textwidth][c]{\begin{tabular}{ccllllll}
     \toprule
    Dataset & $N_\text{cal}$  & Method & Cov. {\tiny $(\gtrsim 0.95$)} & IS ${}^\downarrow$ & $R^2_{SQI}$ ${}^\uparrow$ & $\tau\left(SQI\right)$ ${}^\uparrow$ & $\tau\left(SI\right)$ ${}^\uparrow$\\\midrule
    
    \multirow{6}{*}{1D} & \multirow{3}{*}{500}  & \jprs & $0.96(1)$ & $\mathbf{0.18(1)}$ & $\mathbf{0.87(3)}$ & $0.91(4)$ & $\mathbf{0.35(1)}$ \\
                        &                       & \textsc{MADSplit}  & $0.952(9)$ 	& $\mathbf{0.18(1)}$ 	& $0.80(4)$ 	& $\mathbf{0.96(3)}$ 	& $\mathbf{0.344(9)}$ \\
                        &                       & \textsc{LVD} & $0.982(3)$ 	& $0.21(1)$ 	& $0.61(9)$ 	& $0.86(4)$ 	& $0.33(1)$ \\\cmidrule{2-8}
                        & \multirow{3}{*}{2000} & \jprs & $0.952(2) $ & $\mathbf{0.171(5)}$ & $\mathbf{0.90(2)}$ & $0.96(4)$ & $\mathbf{0.36(1)}$ \\   
                        &                       & \textsc{MADSplit}  & $0.952(9)$ 	& $0.18(1)$ 	& $0.80(4)$ 	& $0.96(3)$ 	& $0.344(9)$ \\                        
                        &                       & \textsc{LVD}  & $0.968(3)$ 	& $\mathbf{0.174(3)}$ 	& $\mathbf{0.92(2)}$ 	& $0.96(3)$ 	& $\mathbf{0.361(6)}$\\\midrule
    \multirow{6}{*}{\shortstack[c]{TDC\\Sol.}} & \multirow{3}{*}{400}  & \jprs & $0.95(1)$ 	& $5.3(5)$ 	& $0.4(1)$ 	& $0.6(1)$ 	& $0.09(1)$ \\
                        &                       & \textsc{MADSplit}  & $0.95(1)$ 	& $5.0(3)$ 	& $\mathbf{0.53(5)}$ 	& $\mathbf{0.71(6)}$ 	& $\mathbf{0.182(3)}$ \\                       
                        &                       & \textsc{LVD}  & $0.958(7)$ 	& $5.2(3)$ 	& $0.3(5)$ 	& $\mathbf{0.7(1)}$ 	& $0.07(5)$ \\\cmidrule{2-8}
                        
                        & \multirow{3}{*}{800} & \jprs & $0.96(1)$ 	& $\mathbf{5.2(4)}$ 	& $0.5(1)$ 	& $0.7(1)$ 	& $0.11(1)$ \\   
                        &                       & \textsc{MADSplit}  & $0.951(6)$ 	& $\mathbf{5.1(2)}$ 	& $\mathbf{0.58(7)}$ 	& $0.71(5)$ 	& $\mathbf{0.179(6)}$ \\                       
                        &                       & \textsc{LVD}  & $0.960(9)$ 	& $5.4(2)$ 	& $0.4(3)$ 	& $0.77(7)$ 	& $0.07(1)$ \\\midrule    
    \multirow{6}{*}{\shortstack[c]{TDC\\Clear.}} & \multirow{3}{*}{60}  & \jprs & $0.95(2)$ 	& $\mathbf{1.9(3)\omag{2}}$ 	& $\mathbf{0.2(3)}$ 	& $\mathbf{0.6(1)}$ 	& $\mathbf{0.37(3)}$  \\
                        &                       & \textsc{MADSplit}   & $0.95(2)$ 	& $\mathbf{1.9(5)\omag{2}}$ 	& $\mathbf{0.2(3)}$ 	& $\mathbf{0.60(9)}$ 	& $\mathbf{0.39(1)}$ \\                       
                        &                       & \textsc{LVD}  & $0.97(2)$ 	& $2.2(2)\omag{2}$ 	& $\ll 0$ 	& $-0.3(6)$ 	& $0.1(2)$\\\cmidrule{2-8}
                        & \multirow{3}{*}{240} & \jprs & $0.96(1)$ 	& $\mathbf{1.65(9)\omag{2}}$ 	& $\mathbf{0.3(2)}$ 	& $\mathbf{0.5(1)}$ 	& $\mathbf{0.39(7)}$  \\   
                        &                       & \textsc{MADSplit}  & $0.96(1)$ 	& $\mathbf{1.64(8)\omag{2}}$ 	& $\mathbf{0.41(5)}$ 	& $\mathbf{0.6(1)}$ 	& $\mathbf{0.41(7)}$ \\
                        &                       & \textsc{LVD} & $0.97(1)$ 	& $2.02(5)\omag{2}$ 	& $\ll 0$ 	& $0.1(2)$ 	& $0.1(1)$\\\midrule
    \multirow{6}{*}{\shortstack[c]{$\alpha$Seq\\CoVID}} & \multirow{3}{*}{2000}  & \jprs & $0.952(6)$ 	& $\mathbf{4.5(1)}$ 	& $\mathbf{0.24(8)}$ 	& $\mathbf{0.5(1)}$ 	& $\mathbf{0.05(1)}$  \\
                        &                       & \textsc{MADSplit}  & $0.952(5)$ 	& $\mathbf{4.6(1)}$ 	& $\mathbf{0.1(2)}$ 	& $0.2(2)$ 	& $0.01(2)$\\                    
                        &                       & \textsc{LVD)}  & $0.994(1)$ 	& $7.6(5)$ &	$\ll 0$ 	& $0.1(1)$ 	& $-0.004(6)$\\\cmidrule{2-8}
                        & \multirow{3}{*}{10000} & \jprs & $0.953(4)$ 	& $\mathbf{4.54(6)}$ 	& $\mathbf{0.28(6)}$ 	& $\mathbf{0.6(1)}$ 	& $\mathbf{0.052(8)}$ \\   
                        &                       & \textsc{MADSplit}  & $0.952(2)$ 	& $\mathbf{4.59(6)}$ 	& $0.0(1)$ 	& $0.2(2)$ 	& $0.01(1)$ \\                
                        &                       & \textsc{LVD}  & $0.973(4)$ 	& $5.49(4)$ 	& $-0.26(4)$ 	& $0.5(1)$ 	& $\mathbf{0.052(8)}$ \\\midrule
    \multirow{6}{*}{MNIST} & \multirow{3}{*}{2000}  & \jprs & $0.952(6)$ 	& $\mathbf{0.62(3)}$ 	& $ \ll 0$ 	& $\mathbf{0.90(5)}$ 	& $0.20(1)$\\
                        &                       & \textsc{MADSplit}  & $0.951(7)$ 	& $\mathbf{0.63(5)}$ 	& $ \ll 0$ 	& $0.83(8)$ 	& $0.17(1)$\\                    
                        &                       & \textsc{LVD)}  & $0.988(4)$ 	& $4.1(5)$ 	& $\mathbf{-0.19(2)}$ 	& $\mathbf{0.86(5)}$ 	& $\mathbf{0.21(1)}$\\\cmidrule{2-8}
                        & \multirow{3}{*}{5000} & \jprs & $0.950(3)$ 	& $\mathbf{0.601(8)}$ 	& $ \ll 0$ 	& $\mathbf{0.92(5)}$ 	& $\mathbf{0.23(1)}$ \\   
                        &                       & \textsc{MADSplit}  & $0.947(4)$ 	& $0.62(1)$ 	& $ \ll 0$ 	& $0.85(7)$ 	& $0.193(6)$ \\                
                        &                       & \textsc{LVD}  & $0.984(2)$ 	& $4.4(2)$ 	& $\mathbf{-0.17(3)}$ 	& $\mathbf{0.92(3)}$ 	& $\mathbf{0.233(9)}$ \\\bottomrule                             
    
    \end{tabular}}
    
    \caption{Numbers in parentheses are the uncertainty on the last significant digit evaluated over 10 random samples of test and calibration data. We report $R^2_{SQI}\ll 0$ if it is significantly lower than~$-2$.}
    \label{tab:big_results}
\end{table}
\section{Limitations and Risks}
\label{sec:limitations}
Our analysis shows our method has great promise for extending the applicability of conformal regression to settings where error rates are variable, its success is dependent on a number of factors whose absence can lead to failure.

\paragraph{Score mean/quantile coupling condition.}
The definition of our conformal scores relies on the hypothesis that the conditional score distribution shifts with $X$ mostly by rescaling. In general, our method's adaptivity will degrade with increasing variance of $q_{1-\alpha}(s|X)/\mathbb{E}(s|X)$, which is exemplified by the results on the MNIST regression task, and discussed in details in~\cref{app:mnist_details}. This ratio cannot be bounded in general, but turns out to be moderate enough in practice for our approach to work in many cases. Alternative constraints can be derived based on concentration inequalities as shown in~\cref{app:concentration}, and might be more suitable empirical assessment.

\paragraph{Non-smooth score-input relationship.}
Our approach based on Nadaraya-Watson estimators of local score distribution means rely on having input representations (combining embeddings and kernels) that change smoothly enough that they can be captured. A poor data representation can break the input-score relationship and break adaptivity. This is an issue with all existing adaptive regression methods and rely on valid design choices.

\paragraph{Risks.}
Coverage guarantees are most commonly formulated in terms of coverage probabilities marginalized on the calibration data, and our method is no exception. While clear to experts, potential downstream users might not realize the variability of conditional coverage, which is paramount to understand for system certification. We suggest that marginal coverage guarantees always be followed by a clear warning of the marginal nature of the guarantee. Software making CP should also provide such warnings or bounds.
For our approach, we've tried to make the current absence of PAC bounds clear, and do so as well in the code to be released with this paper's final version. Given the simple learning algorithm used in the Jackknife+ part of our approach (NW mean estimator), it is likely that we escape the no-go theorem of~\citet{bianTrainingconditionalCoverageDistributionfree2023} and we hope to show in further work that some version of the PAC bounds established in~\citet{barberPredictiveInferenceJackknife2020} apply.

\section{Conclusions}
\label{sec:conclusions}
Our results show that our new approach provides a satisfying solution to the weaknesses of existing post-hoc adaptive conformal regression methods. On the one hand, our use of the Jackknife+ procedure to tune and evaluate the conditional mean of the non-conformity score solves the issue of \textsc{MADSplit} when evaluated on models with significant score distribution shifts between model-training and CP calibration data. On the other hand, while our method does not guarantee perfect local coverage asymptotically unless stringent hypotheses are verified, our empirical results show clear value compared to the diminished statistical power compared to ECDF-reweighting methods, especially in the low-calibration-data regime.

The lack of satisfying tuning criteria for ECDF-based methods also easily leads to infinite PI, even with abundant data, another point in favor of our approach.
Our results indicate that ECDF-based methods tend to have very low tolerance for non-optimal localizers in high-dimensional settings, making further exploration of localizers for ECDF CR critical: these techniques are provably asymptotically optimal adaptive solutions given a good localizer, making them appealing for data-rich domains. Short of that, the comparative robustness of our proposed solution makes it an attractive alternative not only when data is scarce, but also for high-dimensional settings such as deep learning.

It nevertheless remains unclear whether PAC bounds can be established for our method, much like for pre-existing work. 
This is a crucial element for the applicability of any CP method in safety-critical domains, as guarantees marginalized on calibration data cannot be used to certify error rates in specific implementations. 
We have good hope that further investigation of the properties of our method, potentially extending it to a CV+ approach to circumvent the no-go theorem of~\citet{barberConformalPredictionExchangeability2022}, will yield such constraints, further reinforcing our method as a good candidate for uncertainty estimation in regression problems where rigorous uncertainty quantification is essential.

\begin{ack}
We thank Jannis Born, Anna Weber and Aur\'elien P\'elissier for useful discussions.
This work was supported by the Swiss National Science Foundation Grant No.~192128, and by the European Union's Horizon 2020 research and innovation programme under grant agreements No.~101070408 (SustainML) and No.~826121 (iPC).
\end{ack}

\bibliography{main}

\FloatBarrier
\clearpage

\appendix
\appendixpage
\addappheadtotoc
\renewcommand\thefigure{S.\arabic{figure}}
\renewcommand\thetheorem{S.\arabic{theorem}}
\renewcommand\thedefinition{S.\arabic{definition}}
\renewcommand\thepage{S.\arabic{page}}  
\renewcommand\thesection{S.\arabic{section}}  
\renewcommand\thealgorithm{S.\arabic{algorithm}}  

\setcounter{figure}{0}    
\setcounter{theorem}{0}    
\setcounter{definition}{0}    
\setcounter{page}{1}
\FloatBarrier
\section{Algorithms}
\label{app:algo}
In this section, we describe the calibration and PI prediction procedures for two versions of our method in the form of pseudocode. The simplest is~\cref{alg:jplrs}, where the kernel is assumed to be fixed. Indeed we've found in practice that using a KNN kernel with $K=10$ is an efficient and performant approach. We however also define~\cref{alg:kernel_tuning}, where we tune $N$ kernels on calibration data, with an objective motivated by the bound of~\cref{thm:micov}. For high-dimensional data, such as when using embeddings to compute kernel similarities, we use a low-dimensional PCA and compute the sum of the marginal mutual information between the score and principal component. If a PCA is not used, this sum is an upper bound on the multi-dimensional mutual information, but we've also found that using 2-4 principal components captures an overwhelming majority of the data variance for our experiments when using latent space embeddings, and yields good empirical improvement when used for kernel tuning.

\begin{algorithm}
\caption{Jackknife+ rescaled score conformal regression without kernel tuning}\label{alg:jplrs}
\begin{algorithmic}[1]
\Require \\Exchangeable data $\left\{(X_i,y_i) \in \Omega_X\times \mathbb{R} ~|~ {i\in [-T,\dots, N+1]}\right\}$.\\
A learning algorithm $\mathcal A$ mapping a sample of $(X,y)$ pairs to a model $m(X)$.\\
A kernel $K:\Omega_X\times\Omega_X\to \mathbb{R}^+$.\\
A risk threshold $\alpha \in [0,1]$.\\

\Procedure{Predictor Training}{}
\State $m = \mathcal{A}\left(\left\{(X_i,y_i)\right\}_{i\in [-T,\ldots,0]}\right)$
\EndProcedure
\State \textit{All free indices below $(i,j,k)$ span $[1,\dots, N].$}
\Procedure{Calibration}{}
\State \textsc{Set} $p_{ij} = K_{ij}/ \sum_{k\neq i} K_{ik}$
\Comment{$K_{ij} = K(X_i,X_j)$}
\State \textsc{Set} $s^+_i = |y_i-m(X_i)|/ \sum_{k\neq i} p_{ik} s_k$
\EndProcedure
\Procedure{Interval Prediction}{}
\State \textsc{Set} $Q=\lceil (1-\alpha)(N+1) \rceil$
\State \textsc{Set} $p_{i(N+1)} = K_{i(N+1)}/ \sum_{k\neq i,N+1} K_{ik}$
\State \textsc{Set} $\hat{\bar{s}}_{(N+1)i} = \sum_{k\neq i} p_{k(N+1)} s_k$
\State \textsc{Sort} $S = [\hat{\bar{s}}_{(N+1)1} s^+_1,\dots,\hat{\bar{s}}_{(N+1)N} s^+_N]$
\State \textsc{Set} $\Delta y_{N+1} = S[Q]$
\EndProcedure
\Ensure Test prediction interval $[m(X) - \Delta y_{N+1}, m(X)+ \Delta y_{N+1}]$.
\end{algorithmic}
\end{algorithm}

\begin{algorithm}
\caption{Jackknife+ with Kernel Tuning}\label{alg:kernel_tuning}
\begin{algorithmic}[1]
\Procedure{Calibration with Kernel Tuning}{$n_\text{PCA}$, $n_\text{sample}$, $n_\text{scan}$, $\beta_\text{expand}\geq 1$}
\State \textsc{Sample} $n_\text{sample}$ pairs of non-identical training inputs $(X_a,X'_a)$
\State \textsc{Set} $d_\text{min}$, $d_\text{max}$ to the extrema of $||X_a -X'_a ||$
\State \textsc{Set} $\lambda_n$ as $n_\text{scan}$ values  evenly space in logarithmic scale in $[d_\text{min}/\beta_\text{expand}, d_\text{max}\times \beta_\text{expand}]$
\State \textsc{Define} $N$ RBF kernels $K_m$ with length scales $l_m$.
\For {each $\lambda_n$}
\For {$m \in [1,\dots,N]$}
\State \textsc{Set} $l_m=\lambda_n$
\State \textsc{Set} $p_{ij;m} = K_{ij;m}/ \sum_{k\neq i,m} K_{ik;m}$
\Comment{$K_{ij;m} = K_m(X_i,X_j)$}
\State \textsc{Set} $s^+_{i;m} = |y_i-m(X_i)|/ \sum_{k\neq i,m} p_{ik} s_k$
\EndFor
\State \textsc{Set} $\text{mi}_{mn} = \text{MI}\left(\left\{s^+_{i;m}\right\}_{i\neq m}, \left\{X_i\right\}_{i\neq m}\right)$
\EndFor
\State \textsc{Set} $n^*(m) = \underset{n}{\text{argmin}} ~\text{mi}_{mn}$
\State \textsc{Set} $l_m = \lambda_{n^*(m)}$
\State \textsc{Set} $s^+_{i} = s^+_{i;i}$
\EndProcedure
\Procedure{Interval Prediction}{}
\State \textsc{Set} $Q=\lceil (1-\alpha)(N+1) \rceil$.
\State \textsc{Set} $p_{i(N+1)} = K_{i(N+1);i}/ \sum_{k\neq i,N+1} K_{ik;i}$
\State \textsc{Set} $\hat{\bar{s}}_{(N+1)i} = \sum_{k\neq i} p_{k(N+1)} s_k$
\State \textsc{Sort} $S = [\hat{\bar{s}}_{(N+1)1} s^+_1,\dots,\hat{\bar{s}}_{(N+1)N} s^+_N]$
\State \textsc{Set} $\Delta y_{N+1} = S[Q]$
\EndProcedure
\Ensure Test prediction interval $[m(X) - \Delta y_{N+1}, m(X)+ \Delta y_{N+1}]$.
\end{algorithmic}
\end{algorithm}

\section{Dataset and Model Details}
\label{sec:reproducibility}
\subsection{One-Dimensional Regression}

We formulate toy example of regression with label noise in one dimension as follows:
\begin{equation}
\label{eq:app_1d_setup}
\def\arraystretch{2}
\begin{array}{c}
    X\sim \text{U}(0,1), \qquad y = f(X) + \epsilon, \qquad \epsilon  \sim \mathcal{N}\left(0, \lambda \nu(X) \right),\\
    f(X) = f_0 + X^2 \sin\left( k_f X + \phi_f \right), \qquad \nu(X) = \epsilon_0 + \left|\sin\left(k_\epsilon X+ \phi_\epsilon\right)\right|,
\end{array}
\end{equation}

where 
\begin{equation}
    f_0 = 10^{-1},\; k_f = 10,\; \phi_f=1/2,\; \epsilon_0 = 10^{-2},\; k_\epsilon = 2,\; \phi_\epsilon = 3\times 10^{-1},\; \lambda = 10^{-1}.
\end{equation}

For each repetition of the experiment, we sample \textit{i.i.d} pairs of $(X,y)$ and split them into 1000 training points, 10000 test points and a variable number of calibration points. A random-forest regression model is trained, using the default parameters of \textsc{Scikit-Learn}~\texttt{v1.2.2}~\citep{scikit-learn} for our base model, or disabling bootstratpping to produce an overfitting model. For all methods, kernel scores are evaluated on the raw input values $X$

\subsection{Chemical Property Regression on TDC Datasets}

We use two datasets from the Therapeutics Data Commons repository~\citep{huangTherapeuticsDataCommons2021a}, each of which is treated independently.
Both datasets contain samples consisting of pairs of \textsc{SMILES}~\citep{weiningerSMILESChemicalLanguage1988}, descriptions of chemical structure as text sequences, and a target value for the property of interest. The datasets are already divided into training, validation and test samples, however, we merge the validation and test data and re-divide it randomly into calibration and test to allow for enough statistics.
For each task, we fine tune a \textsc{ChemBERTa}~\cite{chithranandaChemBERTaLargeScaleSelfSupervised2020} language model pretrained on \textsc{SMILES} language modelling, provided by \textsc{HuggingFace}.

The first task is to predict drug solubility on data originally produced by~\citet{sorkunAqSolDBCuratedReference2019} and contains 9982 samples, split into 6988 training points and 2994 test points.

The second task consists of estimating the drug microsome clearance (drug elimination rate by the liver) on 1102 samples from~\citet{herseyChEMBLDepositedData2015}, which are split into 772 training points and 330 test points.

Our fine-tuned \textsc{ChemBERTa} models are defined \textsc{HuggingFace} \texttt{AutoModelForSequenceClassification}\footnote{These models can also do regression, despite the name} with \texttt{DeepChem/ChemBERTa-77M-MTR} weights and sequence data is processed with the adapted \texttt{AutoTokenizer}, with maximum sequence lengths set to the maximum sequence length in the training data. The training is performed with the mean-squared-error loss, using a batch size of 64 and a learning rate of $4.0\times 10^{-5}$ over 100 epochs.

Embeddings for kernel scores are computed using the output of the \texttt{classifier.dense} layer of the model, which is the first linear layer of the classification head of the fine-tuned model.

\subsection{AlphaSeq Antibody Affinity Regression}

We use the data from the AlphaSeq survey of 104,972 measurements of the binding affinity of antibody proteins to a SARS-CoV-2 target peptide, paired with amino-acid sequence information for the antibodies. We pre-process the label data by dropping missing measurements and measurements on reference epitopes (keeping only those labelled as \texttt{MIT\_Target}), resulting in 69,297 valid sequence-affinity pairs. The data is divided into a training/validation/test+calibration split of sizes 39466/12517/17314. The test+calibration set is randomly subsampled into a test dataset of size 7314 and a calibration dataset of variable size. 

We use the amino-acid sequences for both the heavy and the light chains of the antibody as inputs to pre-trained, frozen-weight \textsc{AbLang}~\cite{olsenAbLangAntibodyLanguage2022} amino-acid language models adapted to each chain type. The embeddings thus produced are concatenated and mapped to numerical predictions with a two-hidden-layer neural network with layer widths $(128,32)$ and \texttt{ReLU} activations. The final single-output layer does not have an activation function. This model is trained with the \texttt{Adam} optimizer using a learning rate of $1.0\times 10^{-5}$, a batch size of 128 and is regularized with early stopping, monitoring the validation mean-squared error.

Embeddings for kernel scores are the concatenated outputs of the \textsc{AbLang} models.

\subsection{\textsc{MNIST} Regression}
We use the classic \textsc{MNIST} dataset~\citep{lecunMnistDatabaseHandwritten2005} repurposed as a regression task where each digit is labelled by its floating point value. We use data augmentation both at training time and at test time to increase the potential confusion between similar digits, which leads to a non-trivial error structure as we discuss in~\cref{app:mnist_details}.

We use the following randomized distortions, applied sequentially
\begin{itemize}
    \item Gaussian blurring with kernel size $3\times 3$, applied with probability 30\%.
    \item Perspective transformation with scale 0.4, applied  with probability 30\% (\texttt{torchvision} \texttt{RandomPerspective}).
    \item Gaussian noise on each pixel with mean 0 and standard deviation $1/(6+\nu)$ where $\nu\sim \text{U}(0,5)$.
\end{itemize}
The specific choice of transformation is quite arbitrary, but is meant to ensure qualitatively that numbers are nearly always recongnizable to human observers, while making significant distortions common. Transformations are resampled every time an image is used.

\begin{figure}[h!]
    \centering
    \includegraphics[width=0.9\linewidth]{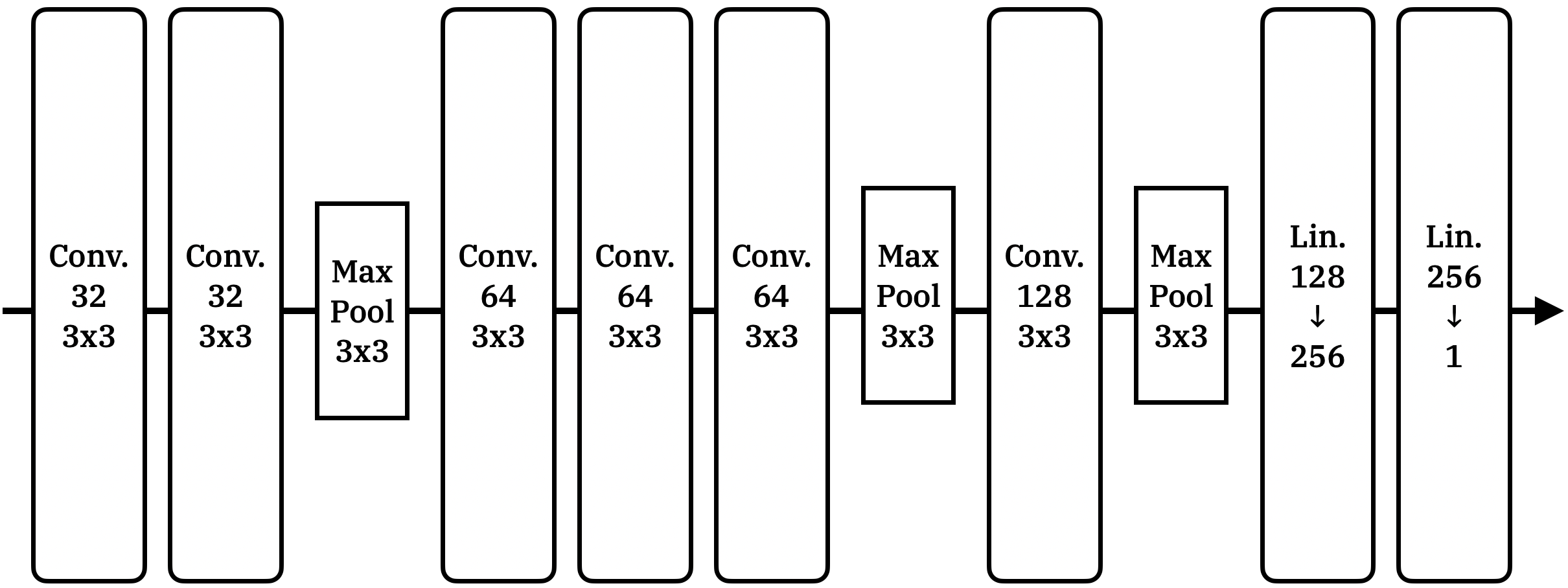}
    \caption{CNN used to perform the \textsc{MNIST} regression task. Each hidden layer is follwed by a batch normalisation layer and a rectified linear unit activation function.}
    \label{fig:arch_mnist}
\end{figure}

We train a convolutional neural network (CNN) on this regression task by optimizing the mean squared error loss with an \textsc{Adam} optimizer with learning rate $5\times 10^{-4}$. The CNN has the architecture described in~\cref{fig:arch_mnist}.

Embeddings for kernel scores are the 256-dimensional outputs of the hidden linear layer.

\section{Additional Experimental Results}
\label{app:more_results}
\subsection{Original Kernel Tuning Strategies}

We present in~\cref{tab:more_results} an extended version of~\cref{tab:big_results} including both our re-tuned kernels for the baselines, as described in~\cref{sec:kernel_comparison} and the original tuning techniques proposed by each method. We report uncertainties as the $1-\sigma$ percentile error, \textit{i.e.} the 67\textsuperscript{th} percentile of absolute deviation from the mean, evaluated with bootstrapping. This is equivalent to reporting the standard deviation if the metrics are normally distributed, which is a weakly-motivated hypothesis given our limited statistics. As we describe in the body of the manuscript, the original tuning strategy proposed for \textsc{LVD} leads to many infinite intervals, sometimes a large majority. Given that it does reach reasonable performance in the 1D example, and that the worse case is \textsc{AlphaSeq}, which uses very high-dimensional embeddings, we conjecture that this reflects the unsuitability of this tuning procedure in high dimensions.

\begin{table}
    \centering
    \makebox[\textwidth][c]{\begin{tabular}{ccllllll}
    \toprule
    Dataset & $N_\text{cal}$  & Method & Cov. {\tiny $(\gtrsim 0.95$)} & IS ${}^\downarrow$ & $R^2_{SQI}$ ${}^\uparrow$ & $\tau\left(SQI\right)$ ${}^\uparrow$ & $\tau\left(SI\right)$ ${}^\uparrow$\\\midrule
    
    \multirow{10}{*}{1D} & \multirow{5}{*}{500}  & \jprs & $0.96(1)$ & $0.18(1)$ & $0.87(3)$ & $0.91(4)$ & $0.35(1)$ \\
                        &                       & \textsc{MADSplit} (KNN) & $0.952(9)$ 	& $0.18(1)$ 	& $0.80(4)$ 	& $0.96(3)$ 	& $0.344(9)$ \\
                        &                       & \textsc{MADSplit} (Median) & $0.95(1)$ 	& $0.19(1)$ &	$ \ll 0$ 	& $0.1(1)$ 	& $0.04(1)$ \\                        
                        &                       & \textsc{LVD} (Tuned)& $0.982(3)$ 	& $0.21(1)$ 	& $0.61(9)$ 	& $0.86(4)$ 	& $0.33(1)$ \\
                        &                       & \textsc{LVD} (Base) & $0.985(2)$ 	& $0.22(1)$ 	& $0.5(1)$ 	& $0.79(9)$ 	& $0.31(2)$ \\\cmidrule{2-8}
                        & \multirow{5}{*}{2000} & \jprs & $0.952(2) $ & $0.171(5)$ & $0.90(2)$ & $0.96(4)$ & $0.36(1)$ \\   
                        &                       & \textsc{MADSplit} & $0.951(3)$ 	& $0.184(2)$ 	& $0.81(2)$ 	& $0.93(6)$ 	& $0.340(7)$ \\
                        &                       & \textsc{MADSplit} (Median) & $0.95(1)$ 	& $0.19(1)$ 	& $\ll 0$ 	& $0.1(1)$ 	& $0.04(1)$ \\                        
                        &                       & \textsc{LVD} (Tuned) & $0.968(3)$ 	& $0.174(3)$ 	& $0.92(2)$ 	& $0.96(3)$ 	& $0.361(6)$\\
                        &                       & \textsc{LVD} (Base) & $0.985(2)$ 	& $0.22(1)$ 	& $0.5(1)$ 	& $0.79(9)$ 	& $0.31(2)$ \\ \midrule
    \multirow{10}{*}{\shortstack[c]{TDC\\Sol.}} & \multirow{5}{*}{400}  & \jprs & $0.95(1)$ 	& $5.3(5)$ 	& $0.4(1)$ 	& $0.6(1)$ 	& $0.09(1)$ \\
                        &                       & \textsc{MADSplit} (KNN) & $0.95(1)$ 	& $5.0(3)$ 	& $0.53(5)$ 	& $0.71(6)$ 	& $0.182(3)$ \\
                        &                       & \textsc{MADSplit} (Median) & $0.95(1)$ 	& $4.6(3)$ 	& $\ll 0$ 	& $0.49(2)$ 	& $0.127(4)$ \\                        
                        &                       & \textsc{LVD} (Tuned) & $0.958(7)$ 	& $5.2(3)$ 	& $0.3(5)$ 	& $0.7(1)$ 	& $0.07(5)$ \\
                        &                       & \textsc{LVD} (Base) & $0.989(2)$ 	& $9(1)$ 	& $-0.8(2)$ 	& $0.6(1)$ 	& $0.06(1)$ \\\cmidrule{2-8}
                        
                        & \multirow{5}{*}{800} & \jprs & $0.96(1)$ 	& $5.2(4)$ 	& $0.5(1)$ 	& $0.7(1)$ 	& $0.11(1)$ \\   
                        &                       & \textsc{MADSplit} (KNN) & $0.951(6)$ 	& $5.1(2)$ 	& $0.58(7)$ 	& $0.71(5)$ 	& $0.179(6)$ \\
                        &                       & \textsc{MADSplit} (Median)& $0.95(1)$ 	& $4.7(2)$ 	& $\ll 0$ 	& $0.51(7)$ 	& $0.124(8)$ \\                        
                        &                       & \textsc{LVD} (Tuned) & $0.960(9)$ 	& $5.4(2)$ 	& $0.4(3)$ 	& $0.77(7)$ 	& $0.07(1)$ \\
                        &                       & \textsc{LVD} (Base) & $0.978(5)$ 	& $7.9(5)$ 	& $-0.1(1)$ 	& $0.8(1)$ 	& $0.058(7)$ \\\midrule    
    \multirow{10}{*}{\shortstack[c]{TDC\\Clear.}} & \multirow{5}{*}{60}  & \jprs & $0.95(2)$ 	& $1.9(3)\omag{2}$ 	& $0.2(3)$ 	& $0.6(1)$ 	& $0.37(3)$  \\
                        &                       & \textsc{MADSplit}  (KNN) & $0.95(2)$ 	& $1.9(5)\omag{2}$ 	& $0.2(3)$ 	& $0.60(9)$ 	& $0.39(1)$ \\
                        &                       & \textsc{MADSplit} (Median) & $0.96(2)$ 	& $2.0(1)\omag{2}$ 	& $\ll 0$ 	& $0.5(1)$ 	& $0.39(1)$ \\                        
                        &                       & \textsc{LVD} (Tuned) & $0.97(2)$ 	& $2.2(2)\omag{2}$ 	& $\ll 0$ 	& $-0.3(6)$ 	& $0.1(2)$\\
                        &                       & \textsc{LVD} (Base& $0.98(1)$ 	& $2.3(2)\omag{2}$ 	& $\ll 0$ 	& $-0.0(3)$ 	& $0.00(5)$ \\\cmidrule{2-8}
                        & \multirow{5}{*}{240} & \jprs & $0.96(1)$ 	& $1.65(9)\omag{2}$ 	& $0.3(2)$ 	& $0.5(1)$ 	& $0.39(7)$  \\   
                        &                       & \textsc{MADSplit} (KNN) & $0.96(1)$ 	& $1.64(8)\omag{2}$ 	& $0.41(5)$ 	& $0.6(1)$ 	& $0.41(7)$ \\
                        &                       & \textsc{MADSplit} (Median)& $0.96(1)$ 	& $1.94(2)\omag{2}$ 	& $\ll 0$ 	& $0.50(7)$ 	& $0.41(5)$ \\
                        &                       & \textsc{LVD} (Tuned)& $0.97(1)$ 	& $2.02(5)\omag{2}$ 	& $\ll 0$ 	& $0.1(2)$ 	& $0.1(1)$\\
                        &                       & \textsc{LVD}  (Base)& $0.995(6)$ 	& $2.57(3)\omag{2}$ 	& $\ll 0$ 	& $-0.1(1)$ 	& $-0.03(4)$ \\\midrule
    \multirow{10}{*}{\shortstack[c]{$\alpha$Seq\\CoVID}} & \multirow{5}{*}{2000}  & \jprs & $0.952(6)$ 	& $4.5(1)$ 	& $0.24(8)$ 	& $0.5(1)$ 	& $0.05(1)$  \\
                        &                       & \textsc{MADSplit} (KNN) & $0.952(5)$ 	& $4.6(1)$ 	& $0.1(2)$ 	& $0.2(2)$ 	& $0.01(2)$\\
                        &                       & \textsc{MADSplit} (Median) & $0.950(5)$ 	& $4.06(7)$ &	$\ll 0$ 	& $0.2(3)$ 	& $0.04(4)$ \\                        
                        &                       & \textsc{LVD)} (Tuned) & $0.994(1)$ 	& $7.6(5)$ &	$\ll 0$ 	& $0.1(1)$ 	& $-0.004(6)$\\
                        &                       & \textsc{LVD} (Base)& $0.9997(3)$ 	& $8.6(8)$& 	$\ll 0$ 	& $0.0(2)$ 	& $-0.003(6)$ \\\cmidrule{2-8}
                        & \multirow{5}{*}{10000} & \jprs & $0.953(4)$ 	& $4.54(6)$ 	& $0.28(6)$ 	& $0.6(1)$ 	& $0.052(8)$ \\   
                        &                       & \textsc{MADSplit} (KNN) & $0.952(2)$ 	& $4.59(6)$ 	& $0.0(1)$ 	& $0.2(2)$ 	& $0.01(1)$ \\
                        &                        & \textsc{MADSplit} (Median) & $0.951(5)$ 	& $4.05(2)$ &	$\ll 0$ 	& $0.1(4)$ 	& $0.01(7)$ \\                        
                        &                       & \textsc{LVD} (Tuned) & $0.973(4)$ 	& $5.49(4)$ 	& $-0.26(4)$ 	& $0.5(1)$ 	& $0.052(8)$ \\
                        &                        & \textsc{LVD} (Base)& $0.99993(6)$ 	& $9.54(3)$ &	$\ll 0$ 	& $0.0(3)$ 	& $-0.001(9)$\\ \bottomrule                         
    
    \end{tabular}}

    \caption{Conformal interval prediction performance metrics on N benchmark datasets. Numbers in parentheses are the uncertainty on the last significant digit evaluated over 10 random samples of test and calibration data.}
    \label{tab:more_results}
\end{table}

\subsection{Jackknife+ Kernel Tuning}
\label{app:kernel_tuning}
For for the sake of limiting computational costs and brevity, we demonstrated the performance of our method with a fixed 10-NN kernel. In this section, we compare this approach with a RBF approach, where the length scale is tuned using~\cref{alg:kernel_tuning}, limiting ourselves to the one-dimensional example.

As we show in~\cref{fig:kernel_tuning_metrics}, a tuned RBF kernel yields comparable results to the KNN approach in the low statistics regime, but achieves better results when more data is available. As we use a Kozachenko-Leonenko-based mutual information estimator, the improved performance in the higher statistics regime is likely due its the asymptotically vanishing bias.

\begin{table}
    \centering
    \makebox[\textwidth][c]{\begin{tabular}{ccllllll}
     \toprule
    Dataset & $N_\text{cal}$  & Method & Cov. {\tiny $(\gtrsim 0.95$)} & IS ${}^\downarrow$ & $R^2_{SQI}$ ${}^\uparrow$ & $\tau\left(SQI\right)$ ${}^\uparrow$ & $\tau\left(SI\right)$ ${}^\uparrow$\\\midrule
    
    \multirow{4}{*}{1D} & \multirow{2}{*}{500}  & \jprs (KNN) & $0.95(1)$ 	& $0.173(9)$ 	& $0.85(4)$ 	& $0.89(7)$ 	& $0.35(1)$ \\
                        &                       & \jprs (tuned) & $0.95(1)$ 	& $0.17(1)$ 	& $0.84(8)$ 	& $0.85(6)$ 	& $0.34(2)$ \\\cmidrule{2-8}
                        & \multirow{2}{*}{2000} & \jprs (KNN) & $0.951(5)$ 	& $0.174(6)$ 	& $0.88(2)$ 	& $0.91(4)$ 	& $0.362(8)$ \\   
                        &                       & \jprs (tuned) & $0.949(3)$ 	& $0.157(4)$ 	& $0.93(1)$ 	& $0.92(4)$ 	& $0.37(1)$\\\midrule
        \end{tabular}}
        \caption{PI metrics evaluated on the 1-dimensional regression task described in~\cref{sec:1d_regression}, comparing our approach with a fixed 10-NN kernel and a RBF kernel tuned with~\cref{alg:kernel_tuning}.}
        \label{fig:kernel_tuning_metrics}
\end{table}

\begin{figure}
    \centering
    \includegraphics[width=0.8\linewidth]{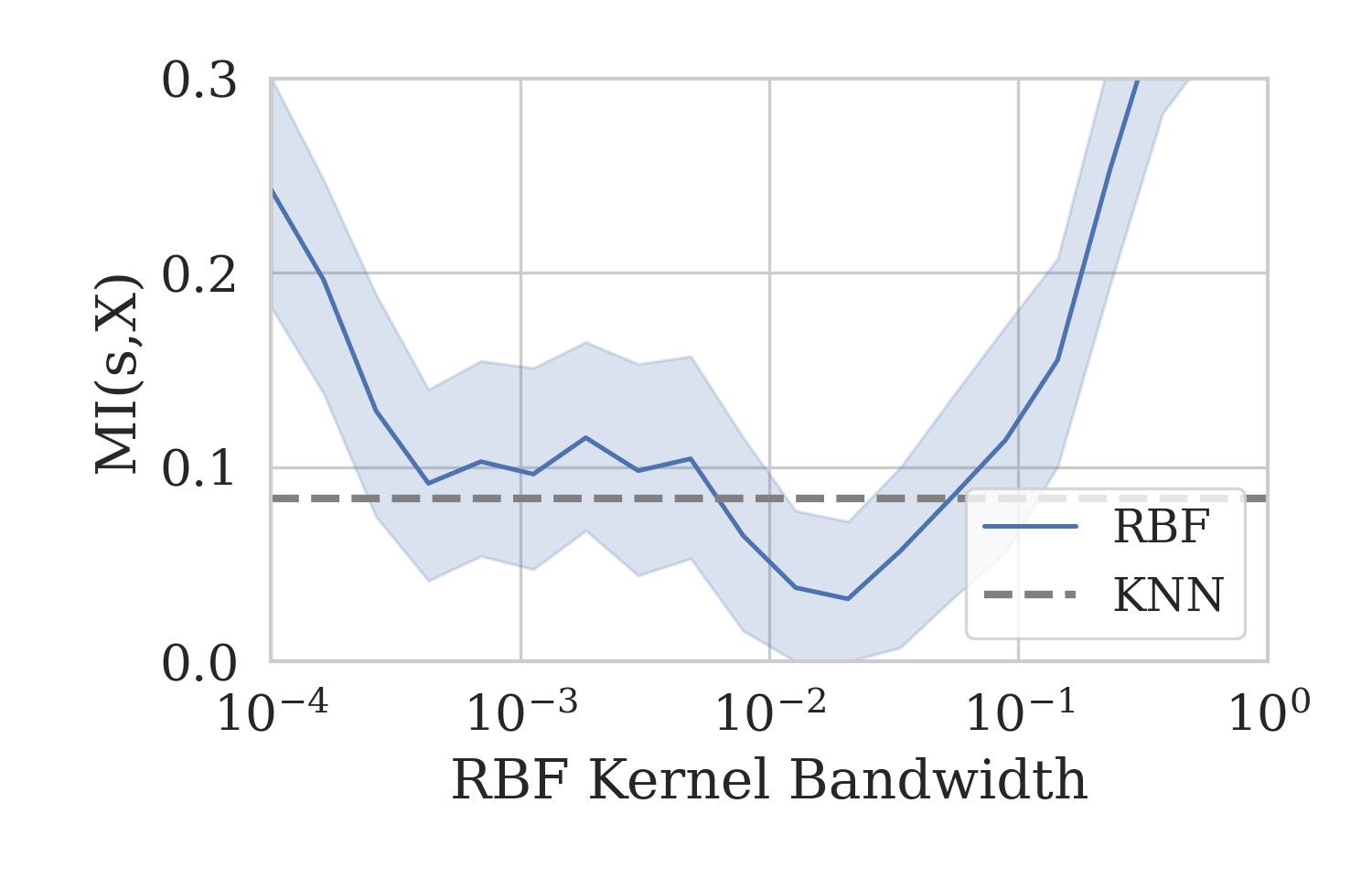}
    \caption{Dependence of the mutual information between $s^+(X,y)$ and $X$ as a function of the kernel length scale, reported as the average over each Jackknife+ split.}
    \label{fig:kernel_tuning_mi}
\end{figure}

This potential for squeezing extra performance is confirmed by investigating the tuning procedure.
In~\cref{fig:kernel_tuning_mi}, we show that the is a finite range where $\textsc{MI}(s^+,X)$ is minimized, and that this minimizer does improve over the same measure evaluated on the 10-NN kernel.

\subsection{AlphaSeq Prediction Details}
\label{app:alphaseq_details}

\begin{figure}
    \centering
    \includegraphics[trim={0.5cm 0 0.9cm 0.9cm},clip, height=0.44\linewidth]{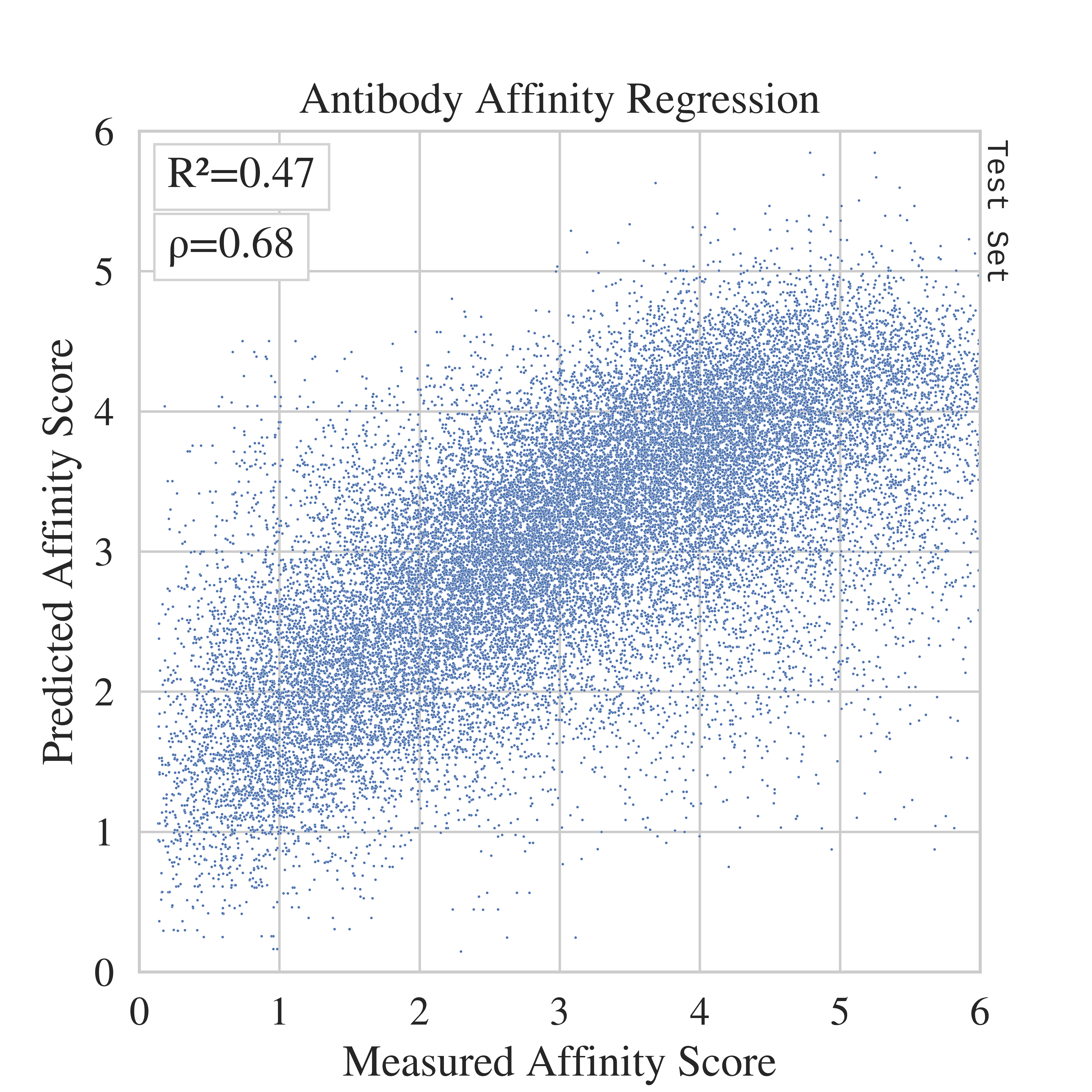}
    \hfill
    \includegraphics[trim={0.9cm 0.4cm 0.6cm 0.5cm},clip, height=0.44\linewidth]{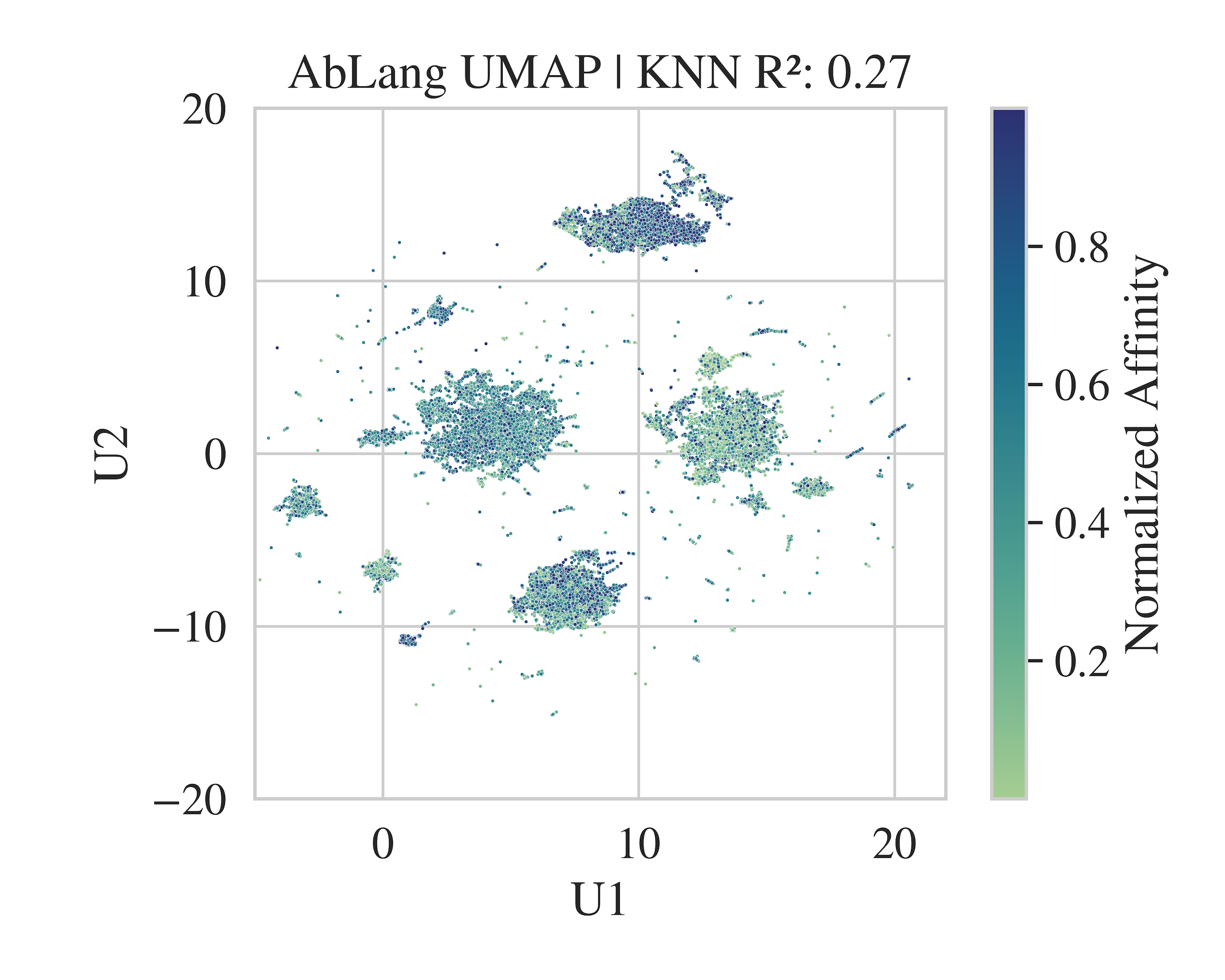}
    \caption{Visualization of our conjectured explanations for the difficulty of \textsc{MADSplit} and \textsc{LVD} on the \textsc{AlphaSeq} dataset. (Left) Regression evaluation plot showing the systematic over and under prediction at the lower and higher edges of the label distribution. (Right) \textsc{UMAP} visualization of the \textsc{AbLang} embeddings used for kernel similarity scores illustrating the scattered structure of the dataset.}
    \label{fig:alphaseq_extras}
\end{figure}

As we discuss in~\cref{sec:exp_many_datasets}, the performance of the baselines is rather poor on the AlphaSeq dataset, which might be especially surprising for \textsc{LVD} due to the large calibration sets. The difficulties of \textsc{MADSplit} can be attributed to the clearly observable overfitting of our trained model, which shows in the form of an S-shape of the label-prediction plot of the test data, as seen in~\cref{fig:alphaseq_extras} (left). We attribute the degraded performance of \textsc{LVD} to the structure of data in the latent space used for similarity measurements: as we show in~\cref{fig:alphaseq_extras} (right), a \textsc{UMAP} of the test data highlights that despite much of the data being concentrated in large clusters, there is a significant number of isolated communities, which are particularly troublesome for \textsc{LVD}. Of course, \textsc{UMAP} representations can obfuscate many features of the data layout, and this explanation is only tentative.

\subsection{Analysis of the MNIST Regression Task Error Distribution}
\label{app:mnist_details}

The \textsc{MNIST} Regression task stands out as the only one where no approach yields even a positive $R^{2}_{SQI}$, despite achieving rather high correlation scores (\cref{tab:big_results}).

\begin{figure}
    \begin{center}
    \includegraphics[width=\linewidth]{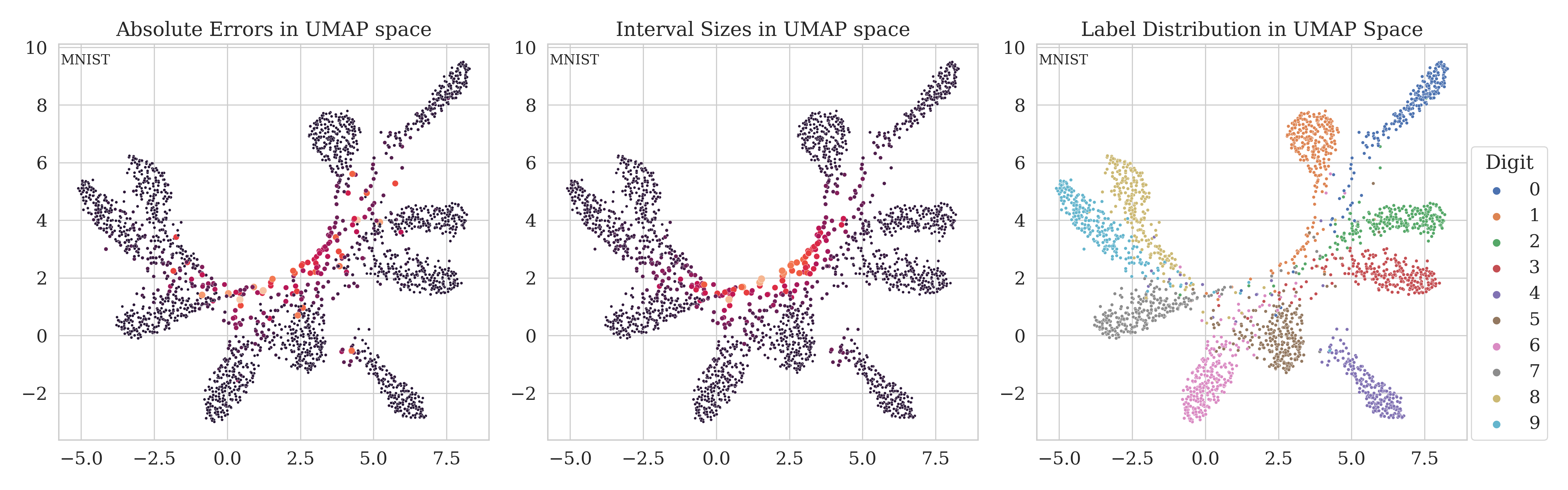}
    \end{center}
    \includegraphics[width=0.95\linewidth]{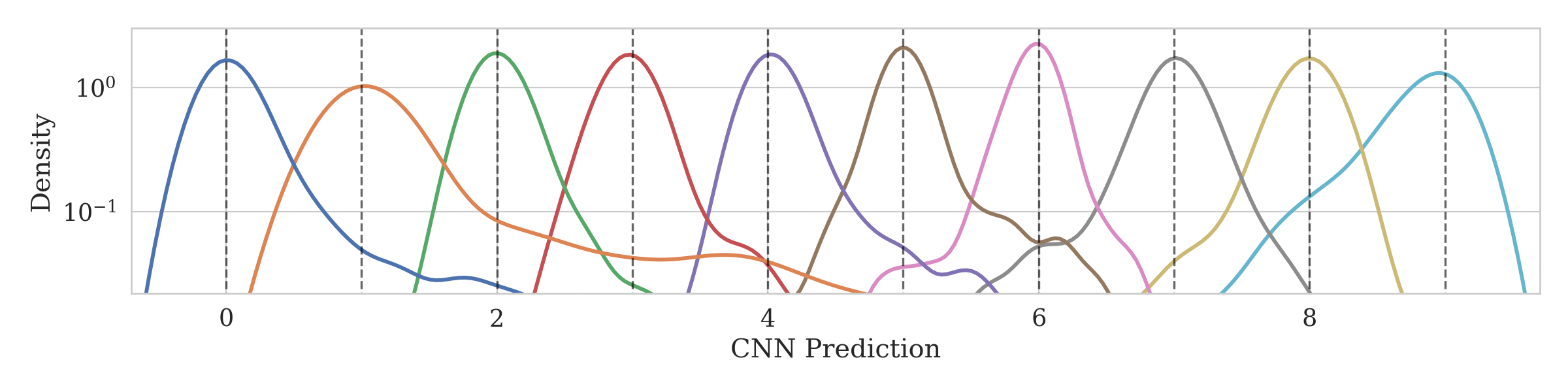}
    \begin{center}
    \caption{(Top) Latent-space structure for the \textsc{MNIST} Regression task interval sizes and labels visualized with UMAP. (Bottom) Prediction distribution per true label.}
    \label{fig:mnist_umap}
    \end{center}
\end{figure}

This observation can be explained by the reason that motivated our choice of this dataset, despite its seemingly contrived nature: the bulk and the tail of the per-label error distributions can be expected to be governed by two independent phenomena. On the one hand, most images are associated with a floating-point label that is distributed around the correct value\footnote{This spread is essentially dictated by the equilibrium reached between minimizing the MSE on clearly-identifiable examples, model capacity, and regularisation.}, as can be seen in~\cref{fig:mnist_umap} (bottom), where the predictions for each digit have similar distributions close to the correct value. On the other hand, this same figure shows that the tail of the error distribution is very class-dependent. This tail structure is dictated by the confusion between digits induced by our test-time data augmentation, which can be observed in latent space, as shown in~\cref{fig:mnist_umap} (top): most of each class is well separated from the others, but each class has a tail overlapping with other classes, inducing high errors. Different digits have different fractions of confusing examples as well as different possible error classes, leading to the observed error dependence. As a result, our hypothesis that the conditional error quantiles is essentially some input-dependent rescaling of the conditional mean is badly broken. The error mean nevertheless captures some dependence of the error on the inputs, but the absolute interval size is thrown off, which is reflected in the high correlation, but low $R^2_{SQI}$.

\section{The Mean Interval Size is not a Sufficient Measure of Adaptivity}
\label{app:mis_limits}
We argue that average interval sizes (IS) are a crude metric for measuring the adaptivity of a PI prediction method. Indeed, like any average-based metric, IS is sensitive to large outliers: a minority population with large errors would have little impact on the PI of a non adaptive method while a dynamic PI prediction would assign larger interval to this sub-population and lead to an increase in IS.

To illustrate this point Let us consider a regression dataset with $N$ elements whose regression errors follow a half normal distribution $\left|\mathcal{N}(0,\sigma)\right|$, where a fraction $\beta$ of points have scale factor $\sigma=\sigma_0$, while the other $(1-\beta)$ have $\sigma=\lambda\sigma_0$.
If $N,\,\lambda >>1$ and $\beta>0.95$, the standard conformal regression intervals based on absolute errors will be dictated by the normal errors and will therefore yield interval sizes close to twice the 95\textsuperscript{th} of the half-normal distribution:
\begin{equation}
    \text{IS}_\text{flat} = 2\times1.96\sigma_0\quad . 
\end{equation}
On the other hand, a perfectly adaptive PI predictor would yield this same $2\times1.96\sigma_0$ interval size for the $\sigma=\sigma_0$ population and $2\times1.96\lambda\sigma_0$ for the others. The average interval size of the "perfect" adaptive PI would therefor be
\begin{equation}
    \text{IS}_\text{perfect} = 2\times 1.96 \sigma_0 \left(\beta + (1-\beta)\lambda \right),
\end{equation}
which will be larger than the flat average interval sizes for large enough $\lambda$.

\begin{figure}[h!]
    \centering
    \includegraphics[width=0.9\linewidth]{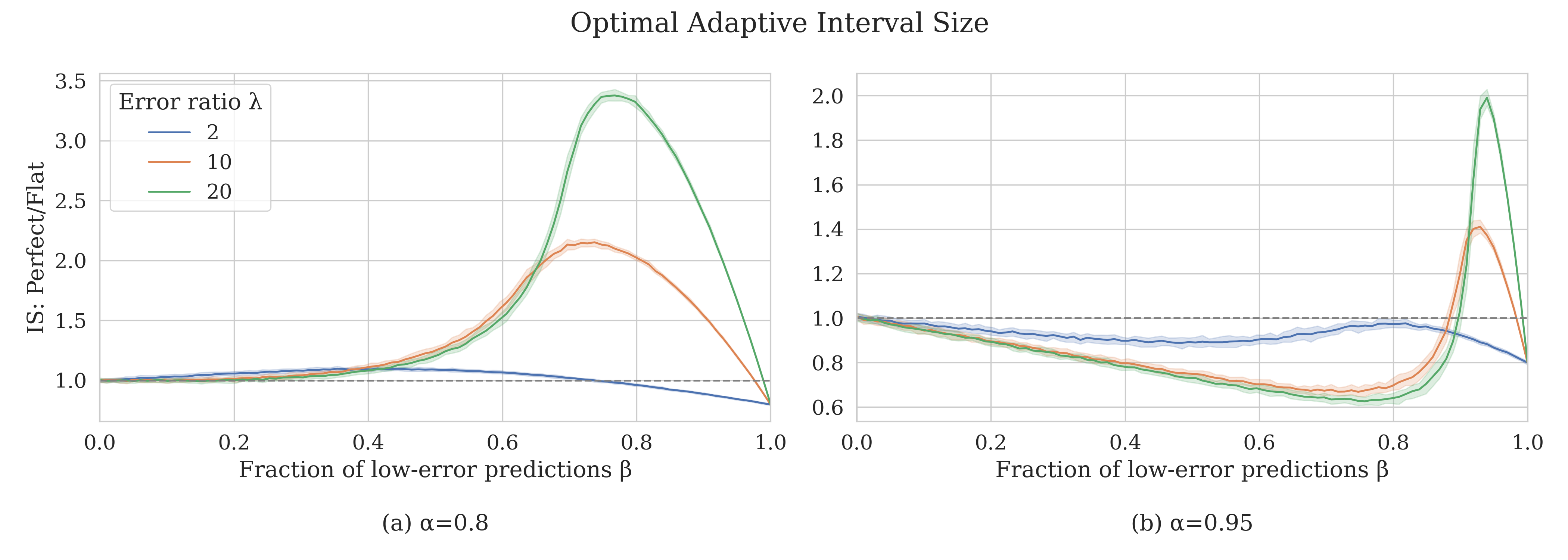}
    \caption{Average interval size ratio between a flat conformal and a perfectly adaptive PI when a errors are normally distributed, with $\sigma=\sigma_0$ for a fraction $\beta$, and $\sigma=\lambda \sigma_0$ for the remaining $1-\beta$.}
    \label{fig:is_ratio}
\end{figure}

We illustrate this point empirically in~\cref{fig:is_ratio}. As expected, the optimally adaptive PI increases when non-coverage risk $\alpha$ is comparable with the fraction of high error examples. This means that IS is inappropriate to use when the error distribution has long tail due to rare difficult examples, which is a common situation in image processing for example~\citep{carliniDistributionDensityTails2019,baldockDeepLearningLens2021,feldmanDoesLearningRequire2021}, or due to skewed heteroscedatic errors.

It is usually desirable to reduce IS at constant coverage, but this reduction is not necessarily a signal of improved adaptivity: the PI predictor might be systematically miscovering high-error minority population, favoring low-error minorities. This observation is what motivates our proposal for more correlation-oriented metrics in~\cref{sec:eval_metrics}, which are sensitive to whether coverage distributes uniformly across high and low error populations.

\section{Formal Global Coverage Guarantees}
Here we show how the conformal intervals defined in \cref{eq:jplus_interval} provides the following theoretical guarantee:

\begin{equation}
P\left(Y_{N+1} \in C^{+\alpha}(X_{N+1})\right) \geq 1-2\alpha.
\end{equation}

The proof is nearly identical to that provided in Section 6 of \citet{barberPredictiveInferenceJackknife2020} and follows in four parts:
\begin{itemize}
\item We define a matrix of score competions between all the $\left\{s^{+}_{ij}\right\}_{i,j\in [1..N+1]}$, whose rows and columns are distributionally permutation-invariant.
\item We use a theorem from Landau to show that there is an upper bound on the number of points that win atypically many competitions
\item We use the distributional permutation invariance of the competition matrix to show that there is an upper bound on the probability that the test point $Z_{N+1}$ is such an ``atypical winner''.
\item We show that $Y_{N+1}\not \in C^{+\alpha}$ implies that $Z_{N+1}$ is an atypical winner, therefore obtaining an upper bound on the probability of this event by contraposition.
\end{itemize}

Let's get through each one

\subsection{Score and competition matrices}

Let us define the following score matrix that has manifest distributional permutation invariance due to the exchangeability of $X_{1},\dots, X_{N+1}$:

\begin{equation}
R = \left(R_{ij}\right) = \left\{
\begin{array}
{l}s^{+}_{ij} \text{ for } i\neq j\in [1..N+1] \\
R_{ii}=\infty 
\end{array} \right.
\end{equation}

From this matrix, be further build a competition matrix

\begin{equation}
A_{ij} = \text{Indicator}\left(R_{ij} > R_{ji}\right).
\end{equation}

Following the original proof, we define the set of ``strange'' points $S(A)$ as those that win abnormally many competitions:

\begin{equation}S(A) = \left\{i\in[1..N+1] \,\Bigg|\: \sum_{j}A_{ij} \geq (1-\alpha)(N+1)\right\}.
\end{equation}

\subsection{Bounding strange points}

There is a finite budget of victories in $A$ since $A_{ij}=1\Leftrightarrow A_{ji}=0$, so there is an upper bound on the size of $S(A)$. This is formalized in a theorem from~\citet{landauDominanceRelationsStructure1953} that implies

\begin{equation}
\left|S(A)\right|\leq 2\alpha(N+1).
\end{equation}

\subsection{From set sizes to probabilities}

The matrix $R$, and therefore $A$ is distributionally permutation invariant, meaning that for any permutation matrix $\Pi$ and any possible matrix value $A_{0}$, $P\left(A=A_{0}\right) = P\left(A=\Pi A_{0} \Pi^{T}\right)$. Permutations on the rows of $A$ correspond to permuting the $X_{i}$ so that

\begin{equation}
P\left(X_{N+1}\in S(A)\right) = P\left(X_{j}\in S\left(\Pi_{j,N+1}A\Pi_{j,N+1}^{T}\right)\right) = P\left(X_{j} \in S(A)\right),
\end{equation}

where $\Pi_{j,N+1}$ is a permutation matrix exchanging rows $j$ and $N+1$. Therefore the probability that $N+1\in S(A)$ is

\begin{equation}
\frac{\left\langle \left| S(A) 
\right|\right\rangle}{N+1}\leq 2\alpha,
\end{equation}

where the inequality follows from the bound on $\left|S(A)\right|$.
\subsubsection{Connecting strange points to coverage}

Let us suppose that $X_{N+1}$ is not covered by its interval $C^{+\alpha}$, this implies that

\begin{equation}
\left|\mu(X_{N+1})-Y_{N+1}\right| > q^{\alpha}\left(\left\{ \sigma_{i,N+1} \frac{\left|\mu(X_{i})-Y_{i}\right|}{\sigma_{N+1,i}} \right\}\right),
\end{equation}

\textbf{i.e.} that for at least $(1-\alpha)(N+1)$ indices $j\in [1..N]$, we have $s^{+}_{(N+1)j} \geq s^{+}_{j(N+1)}$, thus making $X_{N+1}$ a strange point, which has probability bounded by $1-2\alpha$, therefore proving the coverage guarantee we anounced.

\section{Method-Agnostic Results on Local Coverage}
\subsection{Position-independent score distributions guarantee strong input-space local coverage}
\label{app:scoreinputindep}

The intuition behind our localized scores $s^{+}(X,y)$ is that we try to build scores that are not sensistive to local variations of the size of errors so that we can use the whole calibration dataset while getting prediction intervals that are tuned to the local error scale.

This can be formalized as follows: our goal with $s^{+}$ is to ensure that given a random variable $(X,y)\sim \pi(X,y)$, we have $X\perp s^{+}(X,y)$. This is actually a sufficient condition for strong input-space local coverage:

\begin{proposition}
\label{thm:islclemma}
Consider the data $M_{X\times y}$, $\mu$, $s$, $S^{(\alpha)}$ defined in \cref{sec:defs}.\\[0pt]
Let $(X,y)\sim \pi$ be a random variable, then
\begin{equation*}
X \perp s\left(X,y\right) \Rightarrow \alpha-\textsc{iscc}.
\end{equation*}
\end{proposition}

\begin{proof}[Proof of \cref{thm:islclemma}]
\label{proof:islclemma}
Let us remember how the conformal sets $S^{(\alpha)}$ are obtained: we sample a size $n$ calibration set $X_{i},y_{i} \stackbin{\text{i.i.d.}}{\sim}\pi$, define $\hat{q}_{n}^{(\alpha)}$ as the $\left\lceil (1-\alpha)(n+1)\right\rceil / (n+1)\text{-th}$ empirical quantile of the empirical distribution of $\left\{s(X_{i},y_{i})\right\}_{1\leq i \leq n}$ and define $S^{\alpha}(X) = \left\{ \hat{y}\in \Omega_{y} | s(X,\hat{y}) \leq \hat{q}_{n}^{\alpha} \right\}$.\\[0pt]

Having $(X,y) \perp (X_{i}, y_{i})$ and $(X,y)\sim \pi$ ensures the global coverage probability:
\begin{equation}
\mathbb{P}\left(y\in S^{(\alpha)}(X)\right) \geq 1-\alpha.
\end{equation}

Let us assume $X \perp s\left(X,y\right)$. Under our independence assumption, the conditional PDF of $s(X,y)$ verifies $P\left( s(X,y) \Big | X \right) = P\left(s(X,y)\right)$, or
\begin{gather}
\forall \omega_{X}\in \mathcal{F}_{X},\, \omega_{y}\in \mathcal{F}_{y},\, \sigma\in \mathcal{B}\left(\mathbb{R}\right),\nonumber \\
\mathbb{P} \left( s(X,y) \in \sigma_{Xy} \right) = \mathbb{P} \left( s(X,y) \in \sigma | X \right)
\end{gather}

so that $\mathbb{P}\left( s(X,y) \leq q^{(\alpha)}_{n} \Big | X \in \omega_{X} \right) = \mathbb{P}\left(s(X,y) \leq q^{(\alpha)}_{n}\right)$.

Given that $y\in S^{(\alpha)} \Leftrightarrow  s(X,y) \leq q^{(\alpha)}_{n}$ and $\mathbb{P}\left(y\in S^{(\alpha)}(X)\right) \geq 1-\alpha$, we find the desired property holds under our assumptions:
\begin{equation}
\forall \omega_{X} \in \mathcal{F}_{X},\, \mathbb{P}_{X,y\sim \pi}\left(y\in S_{\pi,\mu}^{(\alpha)}\left(X\right) \Big| X\in \omega_{X}\right) \geq 1-\alpha.
\end{equation}
\end{proof}

\subsection{Extending local coverage guarantees to imperfect independence}
\label{app:imperfect}

In practice, the independence of score and inputs will never be realized perfectly. We can nevertheless provide a bound on the local coverage based on their degree of independence, measured as a statistical distance between their joint and product distributions $p_{Xs}(X,s(X,y))$ and $p_{X}(X)\otimes p_{s}(s(X,y))$.

\begin{theorem}
Consider the data $M_{X\times y}$, $\mu$, $s$, $S^{(\alpha)}$ defined in \cref{sec:defs}.

Let $(X,y)\sim \pi$ be a random variable. Let $\textsc{MI}_{Xs}=\textsc{MI}(X,s(X,y))$ bet the mutual information and assume $0<\textsc{MI}_{Xs}<\infty$, then on any $\omega_{X}\in \mathcal{F}_{X}$ such that $0<\mathbb{P}\left(X\in\omega_{X}\right)<\infty$,
\begin{equation}
\mathbb{P}\left( y \in S(X) | X\in \omega_{X}  \right) \geq (1-\alpha) - \frac{\sqrt{1-\exp\left(-\textsc{MI}_{Xs}\right)}}{\mathbb{P}(X\in \Omega(X))}.
\end{equation}
\label{thm:micov}
\end{theorem}

Note that this implies the simpler
\begin{equation}
\mathbb{P}\left( y \in S(X) | X\in \omega_{X}  \right) \geq (1-\alpha) - \frac{\sqrt{\textsc{MI}_{Xs}}}{\mathbb{P}(X\in \Omega(X))},
\end{equation}
which is a pretty tight approximation for small $\textsc{MI}_{Xs}$ (better than $4\%$ for $\textsc{MI}_{Xs}\leq 0.3$) but becomes vacuous faster for large values.

This theorem follows from the Bretagnolle-Huber theorem (see below) and the following lemma:

\begin{lemma}[main technical result]
\label{thm:tvcov}
Let $p_{Xs}$ be the probability density of $(X,s(X,y))$ and $p_X$, $p_s$ the marginal densities of $X$ and $s(X,y)$. If $p_{Xs}(X,s)$ and $p_{X}(X) \otimes p_{s}(s)$ have finite total variation $\delta_{Xs}$, then on any $\omega_{X}\in \mathcal{F}_{X}$,
\begin{align}
&\left|\mathbb{P}\left( y\in S^{\alpha}(X) | X\in \omega_{X}  \right) - (1-\tilde\alpha)\right| \times \mathbb{P}(X\in \omega_{X}). \leq \delta_{Xs}\\
\intertext{where}
&1-\tilde\alpha = \frac{\left\lceil (1-\alpha)(n+1) \right\rceil}{n+1}.
\end{align}

\end{lemma}

\begin{citethm}[Bretagnolle-Huber]

Given two proability distributions $P$ and $Q$ such that $P\ll Q$, then their total variation $\delta(P,Q)$ verifies
\begin{equation}
\delta(P,Q) \leq \sqrt{1-\exp\left(-D_\text{KL}(P||Q) \right)}.
\end{equation}
\end{citethm}
In particular, given two random variables $X,Y$, we have a bound on the total variation between their joint and product distributions expressed in terms of their mutual entropy:
\begin{equation}
\delta\left( p(X,Y), p(X)p(Y) \right) \leq \sqrt{1-\exp\left(-\textsc{MI}(X,Y) \right)}.
\end{equation}

Let us now prove \cref{thm:tvcov}, which is the real meat of the result.

\begin{proof}[Proof of \cref{thm:tvcov}]
Let $\omega_{X}\in \mathcal{F}_{X}$ such that $0<\left|\omega_{X}\right|<\infty$. Let furthermore $q\in [0,1]$ represent the quantile value used to achieve $(1-\alpha)$ global coverage.
We define $\tilde \alpha$ such that $\mathbb{P}\left(s(X,y)\leq q\right)=1-\tilde\alpha$, \textit{i.e.}
\begin{equation}
(1-\tilde\alpha) = \frac{\left\lceil (1-\alpha)(n+1) \right\rceil}{n} \geq (1-\alpha).
\end{equation}
Let us consider the following quantity, writing $s$ as shorthand for $s(X,y)$:
\begin{align}
\Delta(\omega_{X}) &=\left|  \int_{\omega_{X}} dX \int_{0}^{q}ds\, p_{Xs}(X,s)-p_{X}(X)p_{s}(s)\right|\\
\intertext{the integrand is the absolute difference of probabilities over the measurable set $\omega_{X}\times [0,q]$. By definition it is smaller that its supremum over all measurable sets; therefore}
\Delta(\omega_{X})&\leq \delta_{Xs}.\\
\intertext{Furthermore,}
\Delta (\omega_{X})&=\left| \mathbb{P}\left(s\leq q,\phantom{\Big|} X\in \omega_{X}\right) - (1-\tilde\alpha)\mathbb{P}\left(X\in \omega_{X}\right) \right|\\
 &=\left| \mathbb{P}\left(s\leq q \Big| X\in \omega_{X}\right) - (1-\tilde\alpha) \right|\mathbb{P}\left(X\in \omega_{X}\right).
\end{align}
\end{proof}

The bound becomes vacuous on very unlikely sets, which seems unavoidable. This limitation is related to the fact that $\delta$ or $\textsc{MI}$ are global measures and that the contribution from any small set is small. Therefore, a small set can deviate from the mean by a factor that is inversely proportional to its probability. Nevertheless, decreasing the mutual information uniformly decreases the penalty in the local coverage. We have furthermore observed empirically that minimizing the score-input mutual information four our method improves the PI evaluation metrics, even when the bound derived from the minimum is very weak.

\section{Rescaled Scores and Concentration Inequalities}
\label{app:concentration}
In this section, we provide "inspirational" inequalities that result from straigthforward applications of concentration inequalities to mean-rescaled scores. These inequalities cannot be used to put bounds on the local coverage properties of CR methods, but they give hints of why rescaled scores improve local coverage, and of possible directions to explore to obtain valid local bounds.

Throughout this section, we take a \textsc{MADSplit}-like approach in the sense that we assume that we can directly put a threshold on $s(X)/\hat{\bar{s}}(X)$ to obtain PI. Note that we nevertheless assume perfect moment estimators, while finite-sample kernel methods have non-zero bias and variance, so that the inequalities here do not apply for finite samples.

\paragraph{Markov Local Coverage Guarantee}
This is probably the most straightforward inequality: assume that we have obtained a threshold $q^{\alpha}$ by any method. Markov's concentration inequality ($\mathbb{P}(Z>a)\leq \bar{Z}/a$ for $Z>0$.) applied to the random variable $s(X,y)$ with fixed X implies
\begin{equation}
    \mathbb{P}\left( \frac{s(X,y)}{\mathbb{E}_y s(X,y) } \leq q^\alpha | X \right) \geq 1-\frac{1}{q^\alpha},
\end{equation}
which is an actual local coverage guarantee (assuming a perfect conditional mean estimator). Provided it can be extended to an actual mean estimator, this bound might be competitive for error distributions with long tails.

\paragraph{Cantelli's Inequality}
Let us consider Cantelli's inequality:
\begin{equation}
    \mathbb{P}(X- \bar{X} \geq \lambda) \leq \frac{\sigma^2_X}{\sigma^2_X+\lambda^2}.
\end{equation}
Applying this to the random variable $s(X,y)$ with fixed $X$ and $\lambda = (1-\tau) \bar{s}(X)$, it can be rearranged as 

\begin{equation}
    \mathbb{P}\left(\frac{s(X,y)}{\bar{s}(X)} \geq \tau\right) \leq \frac{\sigma^2_s(X)}{\sigma^2_s(X)+(\tau-1)^2 \bar{s}(X)},
\end{equation}
where $\bar{s}(X)$ and $\sigma^2_s(X)$ are the $X$-conditional mean and variance of $s$.

Solving for $\tau$ so that the right-hand-side equals $\alpha$, we find the following inequality

\begin{equation}
    \mathbb{P}\left(  \frac{s}{\bar{s}(X)} \geq 1+\frac{\sigma_s(X)}{\bar s (X)} \sqrt{\frac{1-\alpha}{\alpha}} \right) \leq \alpha.
\end{equation}

This inequality shows that we can reformulate the hypotheses of our method on the $X$-independence of the ratio $q^{1-\alpha}(s|X)/\bar{s}(X)$ for others on $\sigma_s(X)/\bar{s}(X)$. If the latter is $X$-independent, the inequality above shows the existence of a threshold on $s/\bar{s}(X)$ that guarantees local coverage.

\end{document}